\documentclass{singlecol-new}

\usepackage{graphicx, acronym, ifthen, substr, color}
\usepackage[round]{natbib}

\theoremstyle{TH}{
\newtheorem{theorem}{Theorem}[section]

\newtheorem{definition}[theorem]{Definition}

}

\theoremstyle{THhit}{

}

\theoremstyle{THrm}{

}

\makeatletter
\def\theequation{\arabic{equation}}
\makeatother

\newcommand{\added}[1]{#1}
\newcommand{\changed}[1]{#1}

\newcommand{\white}[1]{\color{white}#1\normalcolor}

\newcommand{\be}{\begin{equation}}
\newcommand{\eeq}[1]{\label{#1}\end{equation}}
\newcommand{\bx}{{\bf X}}
\newcommand{\by}{{\bf Y}}
\newcommand{\bone}{{\bf M_{max}}}
\newcommand{\bfif}{{\bf M_{half}}}

\newcommand{\bxi}{\mbox{\boldmath $\xi$}}
\newcommand{\ra}{1, \ldots, n}

\newcommand{\normallinespacing}{\renewcommand{\baselinestretch}{1.5} \normalsize}

\newcommand{\narrowlinespacing}{\renewcommand{\baselinestretch}{1.0} \normalsize}

\newboolean{final}\setboolean{final}{true}

\newboolean{mactex}\setboolean{mactex}{true}

\newboolean{arxiv}\setboolean{arxiv}{false}

\ifthenelse{\boolean{final}}{}{\usepackage{showlabels}\usepackage{citeext}}

\ifthenelse{\boolean{arxiv}}
	{\newcommand{\arxivfootnote}{\footnote}}
	{\def\arxivfootnote#1{}}

\ifthenelse{\boolean{arxiv}}
	{\def\arxivOnly#1{#1}}
	{\def\arxivOnly#1{}}

\ifthenelse{\boolean{arxiv}}
	{\def\arxivNot#1{}}
	{\def\arxivNot#1{#1}}

\ifthenelse{\boolean{arxiv}}
	{}
	{}

\newcommand{\tfigure}[9]
        {
        \IfSubStringInString{!}{#7}{\begin{figure}[#7]}{\begin{figure}[!t]}
        \IfSubStringInString{mm}{#8}{\vspace{#8}}{}
        \centering
        
        \IfSubStringInString{pdf}{#3}
                {
                \execute{cd images; ln -s #2.pdf .#2.gdf}
	      \includegraphics[#1]{images/#2}
                }
                {
                \execute{cd images; ./pdfcrop.sh #2}
                \includegraphics[#1]{images/#2-crop.pdf}
                }

        \vspace{#6}
        \caption[#4]
                {
                \label{#2}
                #4: #5
                }
        \IfSubStringInString{mm}{#9}{\vspace{#9}}{}
        \end{figure}
        }

\newcommand{\Circlesub}[4]
	{
	\ifthenelse{\boolean{mactex}}{}{\immediate\write18{cd images; ./pdfcrop.sh circle#2}}
	\ifthenelse{\boolean{final}}
		{\hspace{#1}\raisebox{#4}{$\includegraphics[scale=0.5, clip=true, trim=0mm 0mm 0mm 0mm]{images/circle#2-crop.pdf}$}\hspace{#3}}
		{\href{file://localhost/Users/g/Desktop/PhDthesis/images/circle#2.graffle}{\hspace{#1}\raisebox{#4}{$\includegraphics[clip=true, trim=0mm 0mm 0mm 0mm]{images/circle#2-crop.pdf}$}\hspace{#3}}}
	}

\newcommand{\squareG}[3]
	{
	\squaresub{#1}{#2}{#3}{-1pt}
	}
	
\newcommand{\squaresub}[4]
	{
	\immediate\write18{cd images; ./pdfcrop.sh square#2}
	\ifthenelse{\boolean{final}}
		{\hspace{#1}\raisebox{#4}{$\includegraphics[scale=0.75,clip=true, trim=0mm 0.25mm 0.25mm 0mm]{images/square#2-crop.pdf}$}\hspace{#3}}
		{\href{file://localhost/Users/g/Desktop/PhDthesis/images/square#2.graffle}{\hspace{#1}\raisebox{#4}{$\includegraphics[scale=0.75,clip=true, trim=0mm 0.25mm 0.25mm 0mm]{images/square#2-crop.pdf}$}\hspace{#3}}}
	}

\newcommand{\execute}[1]{\immediate\write18{#1}}

\definecolor{tred}{RGB}{255,0,0}

\newcommand{\setCap}[2]{#1}
\newcommand{\getCap}[1]{\acl*{#1}}

\acrodef{PCG}{Projected Conjugate Gradient} 
\acrodef{QP}{quadratic programming}
\acrodef{RBF}{Radial-Basis Function}
\acrodef{ABM}{Agent-Based Modelling}
\acrodef{AI}{Artificial Intelligence}
\acrodef{DAI}{Distributed Artificial Intelligence}
\acrodef{API}{Application Programming Interface}
\acrodef{ARF}{p14ARF human tumor-suppressor gene}
\acrodef{B2B}{business-to-business}
\acrodef{BDP}{Biological Design Pattern}
\acrodef{BGS}{Best Guess Solution}
\acrodef{BIC}{Biologically-Inspired Computing}
\acrodef{BML}{Business Modelling Language}
\acrodef{BPEL}{Business Process Execution Language}
\acrodef{BPMN}{Business Process Modelling Notation}
\acrodef{CAS}{Complex Adaptive Systems}
\acrodef{COBOL}{COmmon Business-Oriented Language}
\acrodef{DBE}{Digital Business Ecosystem}
\acrodef{DE}{Digital Ecosystem}
\acrodef{DEC}{distributed evolutionary computing}
\acrodef{DGA}{Distributed genetic algorithms}
\acrodef{DIS}{Distributed Intelligence System}
\acrodef{DNA}{Deoxyribose Nucleic Acid}
\acrodef{DOP}{DBE Open Protocol}
\acrodef{DSS}{Distributed Storage System}
\acrodef{EAP}{Evolving Agent Population}
\acrodef{ebXML}{e-business eXtensible Markup Language}
\acrodef{EC}{Evolutionary Computing}
\acrodef{ECJ}{Evolutionary Computing in Java}
\acrodef{EE}{Evolutionary Environment}
\acrodef{EFL}{Evolutionary Framework for Language}
\acrodef{FLE}{Framework for Language Ecosystems}
\acrodef{EOA}{Ecosystem-Oriented Architecture}
\acrodef{ESS}{evolutionary stable strategy}
\acrodef{EvE}{Evolutionary Environment}
\acrodef{ExE}{Execution Environment}
\acrodef{FCB}{Framework for Computational Biomimicry}
\acrodef{FFF}{Fitness Function Framework}
\acrodef{FL}{Fitness Landscape}
\acrodef{HWU}{Heriot-Watt University}
\acrodef{ICL}{Imperial College London}
\acrodef{ICT}{Information and Communications Technology}
\acrodef{INTEL}{Intel Ireland}
\acrodef{IPA}{International Phonetic Alphabet}
\acrodef{ISUFI}{Istituto Superiore Universitario di Formazione Interdisciplinare}
\acrodef{JDJ}{Java Developer's Journal}
\acrodef{KC}{Kolmogorov-Chaitin}
\acrodef{LAN}{local area network}
\acrodef{LSE}{London School of Economics and Political Science}
\acrodef{MAS}{Multi-Agent System}
\acrodef{MDL}{Minimum Description Length}
\acrodef{MDM2}{murine double minute 2}
\acrodef{MFT}{Mean Field Theory}
\acrodef{MoAS}{Mobile Agent System}
\acrodef{MOF}{Meta Object Facility}
\acrodef{MUH}{migration and usage history}
\acrodef{NIC}{Nature Inspired Computing}
\acrodef{NN}{Neural Network}
\acrodef{NoE}{Network of Excellence}
\acrodef{OMG}{Open Mac Grid}
\acrodef{OPAALS}{Open Philosophies for Associative Autopoietic Digital Ecosystems}
\acrodef{P2P}{peer-to-peer}
\acrodef{P53}{protein 53}
\acrodef{PDA}{Personal Digital Assistant}
\acrodef{QoS}{quality of service}
\acrodef{REST}{REpresentational State Transfer}
\acrodef{RNA}{Deoxyribose Nucleic Acid}
\acrodef{SAE}{Software Agent Ecosystem}
\acrodef{SBML}{Systems Biology Modelling Language}
\acrodef{SBVR}{Semantics of Business Vocabulary and Business Rules}
\acrodef{SDL}{Service Description Language}
\acrodef{SF}{Service Factory}
\acrodef{SIM}{Social Interaction Mechanism}
\acrodef{SM}{Service Manifest}
\acrodef{SME}{Small and Medium sized Enterprise}
\acrodef{SML}{Service Modelling Language}
\acrodef{SMO}{Sequential Minimal Optimisation}
\acrodef{SOA}{Service-Oriented Architecture}
\acrodef{SOAP}{Simple Object Access Protocol}
\acrodef{SOC}{Self-Organised Criticality}
\acrodef{SOLUTA}{SOLUTA.NET}
\acrodef{SOM}{Self-Organising Map}
\acrodef{SSL}{Semantic Service Language}
\acrodef{STU}{Salzburg Technical University}
\acrodef{SUN}{Sun Microsystems}
\acrodef{SVM}{Support Vector Machine}
\acrodef{TM}{Turing Machine}
\acrodef{UBHAM}{University of Birmingham}
\acrodef{UDDI}{Universal Description Discovery and Integration}
\acrodef{UML}{Unified Modelling Language}
\acrodef{URI}{Uniform Resource Identifier}
\acrodef{UTM}{Universal Turing Machine}
\acrodef{VLP}{variable length population}
\acrodef{VLS}{variable length sequences}
\acrodef{vls}{variable length sequence}
\acrodef{WP}{Work-Package}
\acrodef{WSDL}{Web Services Definition Language}
\acrodef{XMI}{XML Metadata Interchange}
\acrodef{XML}{eXtensible Markup Language}
\acrodef{MD5}{Message-Digest algorithm 5}
\acrodef{GA}{genetic algorithm}
\acrodef{GP}{genetic programming}
\acrodef{MASON}{Multi-Agent Simulator Of Neighbourhoods}
\acrodef{Repast}{Recursive Porous Agent Simulation Toolkit}
\acrodef{JCLEC}{Java Computing Library for Evolutionary Computing}
\acrodef{OWL-S}{Web Ontology Language - Service}
\acrodef{EGT}{Evolutionary Game Theory}
\acrodef{RBF}{Radial Basis Functions}
\acrodef{SWS}{Semantic Web Services}
\acrodef{HDD}{Hard Disk Drive}
\acrodef{SSD}{Solid-State Drive}
\acrodef{digEco}{with the agents, the populations, the agent migration for \acl{DEC}, and the environmental selection pressures provided by the user base, then the union of the habitats creates the Digital Ecosystem}
\acrodef{similarCap}{requests are evaluated on separate {islands} (populations), and so adaptation is accelerated by the sharing of solutions between evolving populations (islands), because they are working to solve similar requests (problems).}
\acrodef{picUser}{will formulate queries to the Digital Ecosystem by creating a request as a {semantic description}, like those being used and developed in \acp{SOA}}
\acrodef{picUserReq}{A population is then instantiated in the user's habitat in response to the user's request, seeded from the agents available at their habitat.}
\acrodef{visOrgCap}{number of agents, in total and of each colour, is the same in both populations. \changed{However, the agent population on the left intuitively shows organisation through the uniformity of the colours across the agent-sequences, whereas the population to the right shows little or no organisation.}}
\acrodef{genCap2}{equal the maximum length would be incorrect}
\acrodef{orgCPcap}{are consistent with the intuitive understanding one would have for the self-organisation}
\acrodef{FLsingleCap}{shows the combination space (power set) of the alphabet $D$ against the fitness values from the {selection pressure} (user request)}
\acrodef{FLsingle2Cap}{The agent-sequences of an evolving population will evolve, clustering around the optimal genome}
\acrodef{FLflatCap}{uniformly random}
\acrodef{FLmul2Cap}{Efficiency $E$ will tend to a maximum below one, because the population of sequences consists of more than one cluster, with each having an Efficiency tending to a maximum of one.}
\acrodef{FLmulCap}{The simplest scenario of clusters is {pure clusters}}
\acrodef{popClusShowCap}{The clusters of the population have Efficiency values tending to a maximum of one, compared to the Efficiency of the population as a whole, which is tending to a maximum significantly below one.}
\acrodef{atomCap}{of a set of agents, such that no single agent can functionally replace any agent-sequence, i.e. their functionality is {mutually exclusive} to one another.}
\acrodef{atom2Cap}{the Physical Complexity measure}
\acrodef{popAtomCap}{Efficiency $E$ of the population is a half, whereas the Efficiency $E_{c}$ for populations with clusters is one, because it supports clustering and therefore non-atomicity}
\acrodef{statesCap}{possible evolutionary path through the state-space $I$}
\acrodef{capStates3}{the {selection pressure} of the evolutionary process}
\acrodef{capStates}{driving it towards the {maximal state} of the {maximum macro-state} $M_{max}$}
\acrodef{as3}{with an abstract representation consisting of a set of}
\acrodef{agentSemantic2}{attribute representing a {property} of the {semantic description}, ranging between one and a hundred.}
\acrodef{as4}{Each simulated agent was initialised with a semantic description}
\acrodef{largeVisCap}{The visualisation shows that our Efficiency $E$ accurately measures the self-organised complexity of the two populations.}
\acrodef{graph32cap}{The Efficiency tends to a maximum of one, indicating that the population consists of one cluster}
\acrodef{graph4cap}{around the included best fit curve, quite significantly at the start, and then decreasing as the generations progressed.}
\acrodef{visClustersCap}{agent-sequences were grouped to show the two clusters}
\acrodef{visClusters2Cap}{expected from (\ref{defineCluster}) each cluster had a much higher Physical Complexity and Efficiency compared to the population as a whole. However, the Efficiency $E_{c}$}
\acrodef{visClusters3Cap}{calculated the}
\acrodef{graphCap}{in the {maximum macro-state} $M_{max}$ only after generation 178 and always after generation 482.}
\acrodef{aScap}{With the mutation rate under or equal to 60\%, the evolving agent population showed no instability}
\acrodef{urlCapUnifrom}{The {observed} frequencies of the application (agent-sequence) size mostly matched the {expected} frequencies}
\acrodef{urlCapGaussian}{The {observed} frequencies of the application (agent-sequence) size matched the {expected} frequencies with only minor variations}
\acrodef{urlpower}{The {observed} frequencies of the application (agent-sequence) size matched the {expected} frequencies with some variation}
\acrodef{urvunifromCap}{The {observed} frequencies for the number of agent attributes mostly matched the {expected} frequencies}
\acrodef{urvgaussianCap}{The {observed} frequencies for the number of agent attributes again followed the {expected} frequencies, but there was variation}
\acrodef{urvpowerCap}{The {observed} frequencies for the number of agent attributes also followed the {expected} frequencies, but there was variation}

\begin{document}

\setcounter{page}{1}

\LRH{G Briscoe and P De Wilde}

\RRH{Self-Organisation of Evolving Agent Populations in Digital Ecosystems}

\VOL{x}

\ISSUE{x}

\PAGES{xxxx}

\PUBYEAR{2010}

\subtitle{}

\title{\changed{Self-Organisation of Evolving Agent Populations in Digital Ecosystems}}

\authorA{Gerard Briscoe*}

\affA{Systems Research Group\\ Computer Laboratory\\University of Cambridge\\CB3 0FD, UK\\E-mail gerard.briscoe@cl.cam.ac.uk\\{*}Corresponding author}

\authorB{Philippe De Wilde\vspace{-2mm}}

\affB{Intelligent Systems Lab\\Department of Computer Science\\Heriot Watt University\\EH14 4AS, UK\\E-mail p.de\_wilde@hw.ac.uk}

\begin{abstract}

\added{We investigate the self-organising behaviour of Digital Ecosystems, because a primary motivation for our research is to exploit the self-organising properties of biological ecosystems. We extended a definition for the complexity, grounded in the biological sciences, providing a measure of the information in an organismÕs genome. Next, we extended a definition for the stability, originating from the computer sciences, based upon convergence to an equilibrium distribution. Finally, we investigated a definition for the diversity, relative to the selection pressures provided by the user requests. We conclude with a summary and discussion of the achievements, including the experimental results.}

\end{abstract}

\KEYWORD{\changed{agent; population; self-organisation; complexity; stability; diversity}}

\begin{bio}\\
\added{Gerard Briscoe is a Research Associate at the Systems Research Group of the Computer Laboratory, University of Cambridge, UK, and a Visiting Scholar at Intelligent Systems Lab of the School of Mathematical and Computer Sciences, Heriot-Watt University, UK. Before this, he was a Postdoctoral Researcher at the Department of Media and Communications of the London School of Economics and Political Science, UK. He received his PhD in Electrical and Electronic Engineering from Imperial College London, UK. He worked as a Research Fellow at the MIT Media Lab Europe, after completing his B/MEng in Computing also from Imperial College London. His research interests include sustainable computing, cloud computing, social media and natural computing.}\vs{8}

\noindent \added{Philippe De Wilde is a Professor at the Intelligent Systems Lab, Department of Computer Science, and Head of the School of Mathematical and Computer Sciences, Heriot-Watt University, Edinburgh, United Kingdom. Research interests: stability, scalability and evolution of multi-agent systems; networked populations; coordination mechanisms for populations; group decision making under uncertainty; neural networks, neuro-economics. He tries to discover biological and sociological principles that can improve the design of decision making and of networks. Research Fellow, British Telecom, 1994. Laureate, Royal Academy of Sciences, Letters and Fine Arts of Belgium, 1988. Senior Member of IEEE, Member of IEEE Computational Intelligence Society and Systems, Man and Cybernetics Society, ACM, and British Computer Society. Associate Editor, IEEE Transactions on Systems, Man, and Cybernetics, Part B, Cybernetics.}\vs{8}

\end{bio}

\maketitle

\thispagestyle{headings}

\section{Introduction}

\changed{Digital Ecosystems are distributed adaptive open socio-technical systems, with properties of self-organisation, scalability and sustainability, inspired by natural ecosystems \citep{thesis, caes, dbebkpub, de07oz}, and are emerging as a novel approach to catalysing sustainable regional development driven by \acp{SME}.} Digital Ecosystems aim to help local economic actors become active players in globalisation, valorising their local culture and vocations, and enabling them to interact and create value networks at the global level \citep{dini2008bid, abcdbe}.

\added{Self-organisation is perhaps one of the most desirable features in the systems that we design, and a primary motivation for our research in Digital Ecosystems is the desire to exploit the self-organising properties of biological ecosystems \citep{Levin}, which are thought to be robust, scalable architectures that can automatically solve complex, dynamic problems. Over time a biological ecosystem becomes increasingly self-organised through the process of \emph{ecological succession} \citep{begon96}, driven by the evolutionary self-organisation of the populations within the ecosystem. Analogously, a Digital Ecosystem's increasing self-organisation comes from the agent populations within being evolved to meet the dynamic \emph{selection pressures} created by the requests from the user base. The self-organisation of biological ecosystems is often defined in terms of the \emph{complexity}, \emph{stability}, and \emph{diversity} \citep{king1983cda}, which we will also apply to our Digital Ecosystems.}

\added{It is important for us to be able to understand, model, and define self-organising behaviour, determining macroscopic variables to characterise this self-organising behaviour of the order constructing processes within, the evolving agent populations \citep{phycom, agentStability, acmMedes}. However, existing definitions of self-organisation may not be directly applicable, because evolving agent populations possess properties of both computing systems (e.g. agent systems) as well as biological systems (e.g. population dynamics), and the combination of these properties makes them unique. So, to determine definitions for the self-organising \emph{complexity}, \emph{stability}, and \emph{diversity} we will start by considering our Digital Ecosystems and the available literature on self-organisation, for its general properties, its application to \aclp{MAS} (the dominant technology in Digital Ecosystems), and its application to our evolving agent populations.}

\section{\added{The Digital Ecosystem}}
\added{Our Digital Ecosystem \citep{bionetics, epi} provides a two-level optimisation scheme inspired by natural ecosystems, in which a decentralised peer-to-peer network forms an underlying tier of distributed agents. These agents then feed a second optimisation level based on an evolutionary algorithm that operates locally on single habitats (peers), aiming to find solutions that satisfy locally relevant constraints. The local search is sped up through this twofold process, providing better local optima as the distributed optimisation provides prior sampling of the search space by making use of computations already performed in other peers with similar constraints. So, the Digital Ecosystem supports the automatic combining of numerous agents (which represent services), by their interaction in evolving populations to meet user requests for applications, in a scalable architecture of distributed interconnected habitats. The sharing of agents between habitats ensures the system is scalable, while maintaining a high evolutionary specialisation for each user. The network of interconnected habitats is equivalent to the \emph{abiotic} environment of biological ecosystems; combined \setCap{with the agents, the populations, the agent migration for \acl{DEC}, and the environmental selection pressures provided by the user base, then the union of the habitats creates the Digital Ecosystem}{digEco}, which is summarised in Figure \ref{architecture2}. The continuous and varying user requests for applications provide a dynamic evolutionary pressure on the applications (agent-sequences), which have to evolve to better fulfil those user requests, and without which there would be no driving force to the evolutionary self-organisation of the Digital Ecosystem.}

\tfigure{scale=0.6}{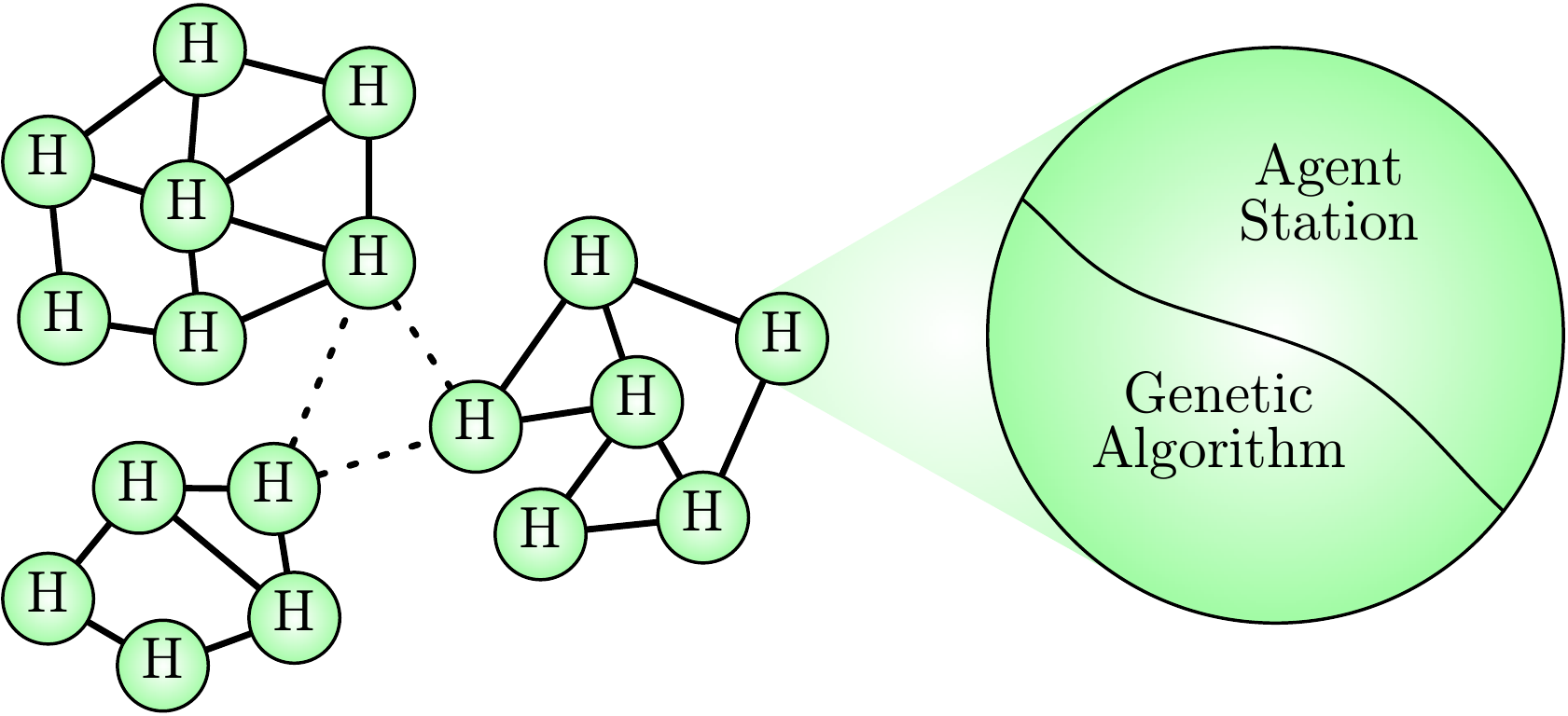}{graffle}{\added{Digital Ecosystem}}{\added{Optimisation architecture in which agents (representing services) travel along the P2P connections; in every node (habitat) local optimisation is performed through an evolutionary algorithm, where the search space is determined by the agents present at the node.}}{0mm}{}{}{}

\added{If we consider an example user base for the Digital Ecosystem, the use of \acp{SOA} in its definition means that \acf{B2B} interaction scenarios lend themselves to being a potential user base for Digital Ecosystems. So, we can consider a \emph{business ecosystem} of \acf{SME} networks \citep{moore1996}, as a specific class of examples for \ac{B2B} interaction scenarios; and in which the \ac{SME} users are requesting and providing software services, represented as agents in the Digital Ecosystem, to fulfil the needs of their business processes, creating a Digital Business Ecosystem as shown in Figure \ref{DBE}. \acp{SOA} promise to provide potentially huge numbers of services that programmers can combine, via the standardised interfaces, to create increasingly more sophisticated and distributed applications. The Digital Ecosystem extends this concept with the automatic combining of available and applicable services, represented by agents, in a scalable architecture, to meet user requests for applications. These agents will recombine and evolve over time, constantly seeking to improve their effectiveness for the user base. From the SME users' point of view the Digital Ecosystem provides a network infrastructure where connected enterprises can advertise and search for services (real-world or software only), putting a particular emphasis on the composability of loosely coupled services and their optimisation to local and regional, needs and conditions. To support these SME users the Digital Ecosystem is satisfying the companies' business requirements by finding the most suitable services or combination of services (applications) available in the network. An application (composition of services) is defined be an agent-sequence in the habitat network that can move from one peer (company) to another, being hosted only in those where it is most useful in satisfying the \ac{SME} users' business needs.}

\added{The agents consist of an \emph{executable component} and an \emph{ontological description}. So, the Digital Ecosystem can be considered a \ac{MAS} which uses \emph{distributed evolutionary computing} to combine suitable agents in order to meet user requests for applications.}

\added{The landscape, in energy-centric biological ecosystems, defines the connectivity between habitats. Connectivity of nodes in the digital world is generally not defined by geography or spatial proximity, but by information or semantic proximity. For example, connectivity in a peer-to-peer network is based primarily on bandwidth and information content, and not geography. The island-models of \acl{DEC} use an information-centric model for the connectivity of nodes (\emph{islands}) \citep{lin1994cgp}. However, because it is generally defined for one-time use (to evolve a solution to one problem and then stop) it usually has a fixed connectivity between the nodes, and therefore a fixed topology. So, supporting evolution in the Digital Ecosystem, with a multi-objective \emph{selection pressure} (fitness landscape with many peaks), requires a re-configurable network topology, such that habitat connectivity can be dynamically adapted based on the observed migration paths of the agents between the users within the habitat network. Based on the island-models of \acl{DEC} \citep{lin1994cgp}, each connection between the habitats is bi-directional and there is a probability associated with moving in either direction across the connection, with the connection probabilities affecting the rate of migration of the agents. However, additionally, the connection probabilities will be updated by the success or failure of agent migration using the concept of Hebbian learning: the habitats which do not successfully exchange agents will become less strongly connected, and the habitats which do successfully exchange agents will achieve stronger connections. This leads to a topology that adapts over time, resulting in a network that supports and resembles the connectivity of the user base. If we consider a \emph{business ecosystem}, network of \acp{SME}, as an example user base; such business networks are typically small-world networks \citep{white2002nst}. They have many strongly connected clusters (communities), called \emph{sub-networks} (quasi-complete graphs), with a few connections between these clusters (communities) \citep{swn1}. Graphs with this topology have a very high clustering coefficient and small characteristic path lengths. So, the Digital Ecosystem will take on a topology similar to that of the user base, as shown in Figure \ref{DBE}.}

\tfigure{scale=0.8}{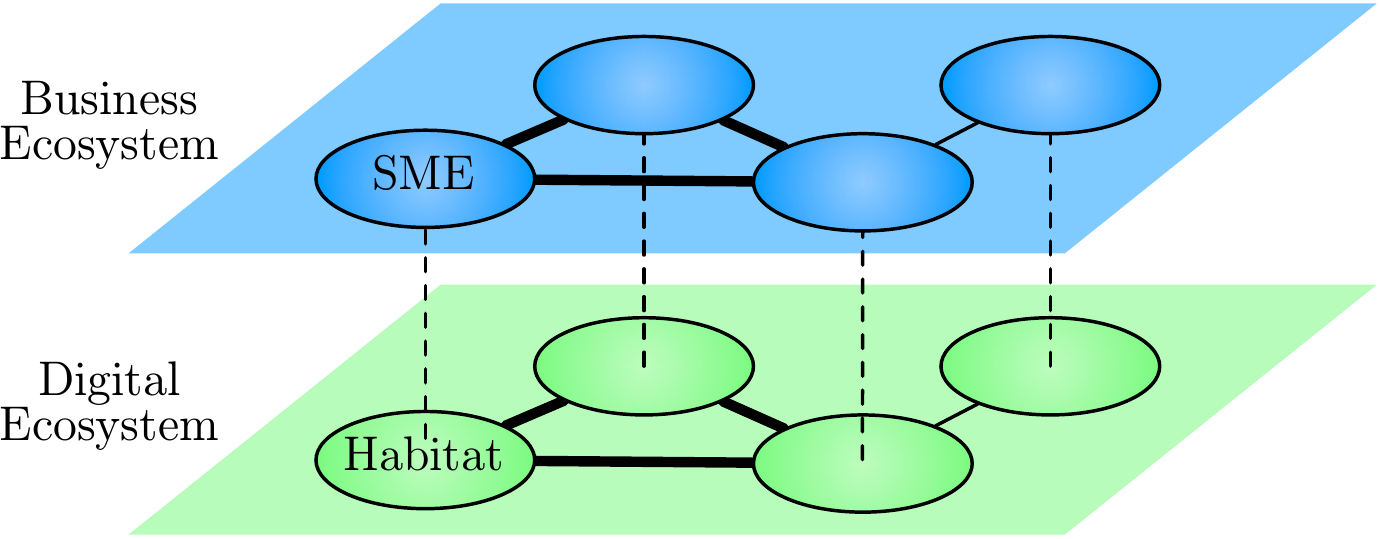}{graffle}{\added{Digital Business Ecosystem}}{\added{Business ecosystem, network of \acp{SME} \citep{moore1996}, using the Digital Ecosystem. The habitat clustering will therefore be parallel to the business sector communities.}}{0mm}{!b}{}{}

\added{The novelty of our approach comes from the evolving populations being created in response to \emph{similar} requests. So whereas in the island-models of \acl{DEC} there are multiple evolving populations in response to one request \citep{lin1994cgp}, here there are multiple evolving populations in response to \emph{similar} requests. In our Digital Ecosystems different \setCap{requests are evaluated on separate \emph{islands} (populations), and so adaptation is accelerated by the sharing of solutions between evolving populations (islands), because they are working to solve similar requests (problems).}{similarCap}}

\added{The users \setCap{will formulate queries to the Digital Ecosystem by creating a request as a \emph{semantic description}, like those being used and developed in \acp{SOA}}{picUser}, specifying an application they desire and submitting it to their local peer (habitat). This description defines a metric for evaluating the \emph{fitness} of a composition of agents, as a distance function between the \emph{semantic description} of the request and the agents' \emph{ontological descriptions}. \setCap{A population is then instantiated in the user's habitat in response to the user's request, seeded from the agents available at their habitat.}{picUserReq} This allows the evolutionary optimisation to be accelerated in the following three ways: first, the habitat network provides a subset of the agents available globally, which is localised to the specific user it represents; second, making use of applications (agent-sequences) previously evolved in response to the user's earlier requests; and third, taking advantage of relevant applications evolved elsewhere in response to similar requests by other users. The population then proceeds to evolve the optimal application (agent-sequence) that fulfils the user request, and as the agents are the base unit for evolution, it searches the available agent combination space. For an evolved agent-sequence (application) that is executed by the user, it then migrates to other peers (habitats) becoming hosted where it is useful, to combine with other agents in other populations to assist in responding to other user requests for applications.}

\section{Self-Organisation}

\changed{Self-organisation has been around since the late 1940s \citep{ashby}, but has escaped general formalisation despite many attempts \citep{nicolis, kohonen1989soa}. There have instead been many notions and definitions of self-organisation, useful within their different contexts \citep{heylighen2002sso}. They have come from cybernetics \citep{ashby, beer1966dac, heylighen2001cas}, thermodynamics \citep{nicolis}, mathematics \citep{lendaris1964dso}, information theory \citep{shalizi2001cac}, synergetics \citep{haken1977sin}, and other domains \citep{lehn1990psc}. The term \emph{self-organising} is widely used, but there is no generally accepted meaning, as the abundance of definitions would suggest. Therefore, the philosophy of self-organisation is complicated, because organisation has different meanings to different people. So, we would argue that any definition of self-organisation is context dependent, in the same way that a choice of statistical measure is dependent on the data being analysed.}

Proposing a definition for self-organisation faces the \emph{cybernetics} problem of defining \emph{system}, the \emph{cognitive} problem of \emph{perspective}, the \emph{philosophical} problem of defining \emph{self}, and the \emph{context} dependent problem of defining \emph{organisation} \citep{gershenson}.

\changed{The \emph{system} in this context is an \emph{evolving agent population}, with the replication of individuals from one generation to the next, the recombination of the individuals, and a selection pressure providing a differential fitness between the individuals, which is behaviour common to any evolving population \citep{begon96}.}

\emph{Perspective} can be defined as the perception of the observer in perceiving the self-organisation of a system \citep{ashby1962pso, beer1966dac}, matching the intuitive definition of \emph{I will know it when I see it} \citep{shalizi}, which despite making formalisation difficult shows that organisation is \emph{perspective} dependent (i.e. relative to the \emph{context} in which it occurs). In the \emph{context} of an evolutionary system, the observer does not exist in the traditional sense, but is the \emph{selection pressure} imposed by the environment, which \emph{selects} individuals of the population over others based on their \emph{observable} \emph{fitness}. Therefore, consistent with the theoretical biology \citep{begon96}, in an evolutionary system the self-organisation of its population is from the \emph{perspective} of its environment. 

Whether a system is \emph{self}-organising or being organised depends on whether the process causing the organisation is an internal component of the system under consideration. \changed{This intuitively makes sense, and therefore requires one to define the boundaries of the system being considered to determine if the force causing the organisation is internal or external to the system.} For an evolving population the force leading to its organisation is the \emph{selection pressure} acting upon it \citep{begon96}, which is formed by the environment of the population's existence and competition between the individuals of the population \citep{begon96}. As these are internal components of an evolving agent population \citep{begon96}, it is a self-organising system.

Now that we have defined, for an evolving agent population, the \emph{system} for which its \emph{organisation} is \emph{context} dependent, the \emph{perspective} to which it is relative, and the \emph{self} by which it is caused, a definition for its \emph{self-organisation} can be considered. \changed{The \emph{context}, an evolving agent population in its environment, lacks a 2D or 3D metric space, so it is necessary to consider a visualisation in a more abstract form.} We will let a single square, \squareG{-5mm}{White}{-3mm}, represent an agent, with colours to represent different agents. Agent-sequences will therefore be represented by a sequence of coloured squares, \squareG{-3.5mm}{RedGreenBlue}{-2mm}, with a population consisting of multiple agent-sequences, as shown in Figure \ref{sampleAgentPopulation}.

\tfigure{scale=1.0}{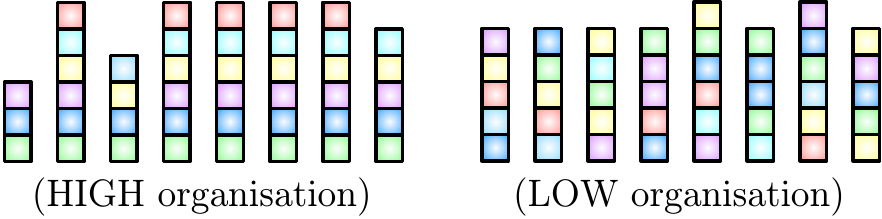}{graffle}{Visualisation of Self-Organisation in Evolving Agent Populations}{The \getCap{visOrgCap}}{0mm}{!b}{}{}

In Figure \ref{sampleAgentPopulation} the \setCap{number of agents, in total and of each colour, is the same in both populations. \changed{However, the agent population on the left intuitively shows organisation through the uniformity of the colours across the agent-sequences, whereas the population to the right shows little or no organisation.}}{visOrgCap} \added{Following biological ecosystems, which defines self-organisation in terms of the \emph{complexity}, \emph{stability}, and \emph{diversity} relative to the \emph{perspective} of the \emph{selection pressure} \citep{king1983cda}: the self-organised \emph{complexity} of the system is the creation of coherent patterns and structures from the agents, the self-organised \emph{stability} of the system is the resulting stability or instability that emerges over time in these coherent patterns and structures, and the self-organised \emph{diversity} of the system is the optimal variability within these coherent patterns and structures.}

\subsection{\added{Definitions of Self-Organisation}}

\added{Many alternative definitions have been proposed for self-organisation within populations and agent systems, with each defining what property or properties demonstrate self-organisation. So, we will now consider the most applicable alternatives for their suitability in defining the self-organised \emph{complexity}, \emph{stability}, and \emph{diversity} of an evolving agent population.}

\added{One possibility would be the $\in$-machine definition of evolving populations, which models the emergence of organisation in pre-biotic evolutionary systems \citep{crutchfield2006}. An $\in$-machine consists of a set of causal states and transitions between them, with symbols of an alphabet labelling the transitions and consisting of two parts: an input symbol that determines which transition to take from a state, and an output symbol which is emitted on taking that transition \citep{crutchfield2006}. $\in$-machines have several key properties \citep{crutchfieldYoung1989}: all their recurrent states form a single, strongly connected component, their transitions are deterministic in the specific sense that a causal state with the edge symbol-pair determines the successor state, and an $\in$-machine is the smallest causal representation of the transformation it implements. The $\in$-machine definition of self-organisation also identifies the forms of \emph{complexity}, \emph{stability}, and \emph{diversity} \citep{crutchfield2006}, but with definitions focused on pre-biotic evolutionary systems, i.e. the \emph{primordial soup} of \emph{chemical replicators} from the \emph{origin of life} \citep{rasmussen2004}. \emph{Complexity} is defined as a form of structural-complexity, measuring the state-machine-based information content of the $\in$-machine individuals of a population \citep{crutchfield2006}. \emph{Stability} is defined as a meta-machine, a set (composition) of $\in$-machines, that can be regarded as an autonomous and self-replicating entity \citep{crutchfield2006}. \emph{Diversity} is defined, using an interaction network, as the variability of interaction in a population \citep{crutchfield2006}. So, while these definitions of self-organisation are compatible at the higher more abstract level, i.e. in the forms of self-organisation present, the deeper definitions of these forms are not applicable because they are context dependent. As we explained in the previous subsection, definitions of self-organisation are context dependent, and so the context of pre-biotic evolutionary systems, to which the $\in$-machine self-organisation applies, is very different to the context of an evolving agent population from our Digital Ecosystem. Evolving agent populations are defined from \aclp{EOA}, which have evolutionarily surpassed the context of pre-biotic evolutionary systems, shown by the necessity of our consideration of the later evolutionary stage of \emph{ecological succession} \citep{begon96} \citep{thesis}.}

\added{The \emph{\acl{MDL}} principle \citep{barron} could be applied to the \emph{executable components} or \emph{semantic descriptions} of the agent-sequences of a population, with the best model, among a collection of tentatively suggested ones, being the one that provides the smallest \emph{stochastic complexity}. However, the \acl{MDL} principle does not define how to select the family of model classes to be applied for determining the stochastic complexity \citep{hansen2001msa}. This problem of model selection is well known and cannot be adequately formalised, and so in practise selection is based on human judgement and prior knowledge of the kinds of models previously chosen \citep{hansen2001msa}. Therefore, while models could be chosen to represent the self-organised \emph{complexity}, and possibly even the \emph{diversity}, there is no procedural method for determining these models, because \emph{subjective} human intervention is required for model selection on a case-by-case basis.}

\added{The \emph{Pr{\"u}gel-Bennett Shapiro formalism} models the evolutionary dynamics of a population of sequences, using techniques from statistical mechanics and focuses on replica symmetry \citep{prugel}. The individual sequences are not considered directly, but in terms of the statistical properties of the population, using a macroscopic level of description with specific statistical properties to characterise the population, that are called \emph{macroscopics}. A macroscopic formulation of an evolving population reduces the huge number of degrees of freedom to the dynamics of a few quantities, because a non-linear system of a few degrees of freedom can be readily solved or numerically iterated \citep{prugel}. However, since a macroscopic description disregards a significant amount of information, subjective human insight is essential so that the appropriate macroscopics are chosen \citep{shapiro2001smt}. So, while macroscopics could be chosen to represent the self-organised \emph{complexity}, \emph{stability}, and \emph{diversity}, there is no procedural method for determining these macroscopics, because \emph{subjective} human insight is required for macroscopic selection on a case-by-case basis.}

\added{\emph{\acl{KC}} complexity defines the complexity of binary sequences by the smallest possible \acl{UTM}, algorithm (programme and input) that produces the sequence \citep{li1997ikc}. A sequence is said to be \emph{regular} if the algorithm necessary to produce it on a \acl{UTM} is shorter than the sequence itself \citep{li1997ikc}. A \emph{regular} sequence is said to be \emph{compressible}, whereas its compression, into the most succinct \acl{UTM} possible, is said to be \emph{incompressible} as it cannot be reduced any further in length \citep{li1997ikc}. A \emph{random} sequence is said to be \emph{incompressible}, because the \acl{UTM} to represent it cannot be shorter than the random sequence itself \citep{li1997ikc}. This intuitively makes sense for algorithmic complexity, because algorithmically regular sequences require a shorter programme to produce them. So, when measuring a population of sequences, the \acl{KC} complexity would be the shortest \acl{UTM} to produce the entire population of sequences. However, Chaitin himself has considered the application of \acl{KC} complexity to evolutionary systems, and realised that although \acl{KC} complexity represents a satisfactory definition of randomness in algorithmic information theory, it is not so useful in biology \citep{chaitin}. For evolving agent populations the problem manifests itself most significantly when the agents are randomly distributed within the agent-sequences of the population, having maximum \acl{KC} complexity, instead of the complexity it ought to have of zero. This property makes \acl{KC} complexity unsuitable as a definition for the self-organised \emph{complexity} of an evolving agent population.}

\added{A definition called \emph{Physical Complexity} can be estimated for a population of sequences, calculated from the difference between the maximal entropy of the population, and the actual entropy of the population when in its environment \citep{adami20002}. This Physical Complexity, based on Shannon's entropy of information, measures the information in the population about its environment, and therefore is conditional on its environment. It can be estimated by counting the number of loci that are fixed for the sequences of a population \citep{adami1998ial}. Physical Complexity would therefore be suitable as a definition of the self-organised \emph{complexity}. However, a possible limitation is that Physical Complexity is currently only formulated for populations of sequences with the same length.}

\added{\emph{\acl{SOC}} in evolution is defined as a punctuated equilibrium in which the population's \emph{critical state} occurs when the fitness of the individuals is uniform, and for which an \emph{avalanche}, caused by the appearance and spread of advantageous mutations within the population, temporarily disrupts the uniformity of individual fitness across the population \citep{bak1988soc}. Whether an evolutionary process displays \acl{SOC} remains unclear. There are those who claim that \acl{SOC} is demonstrated by the available fossil data \citep{sneppen}, with a power law distribution on the lifetimes of genera drawn from fossil records, and by artificial life simulations \citep{adami1995}, again with a power law distribution on the lifetimes of competing species. However, there are those who feel that the fossil data is inconclusive, and that the artificial life simulations do not show \acl{SOC}, because the key power law behaviour in both can be generated by models without \acl{SOC} \citep{newman1996soc}. Also, the \acl{SOC} does not define the resulting self-organised \emph{stability} of the population, only the organisation of the events (avalanches) that occur in the population over time.}

\added{\emph{\acl{EGT}} \citep{weibull1995egt} is the application of models inspired from \emph{population genetics} to the area of \emph{game theory}, which differs from \emph{classical game theory} \citep{fudenberg1991gt} by focusing on the dynamics of strategy change more than the properties of individual strategies. In \acl{EGT}, agents of a population play a game, but instead of optimising over strategic alternatives, they inherit a fixed strategy and then replicate depending on the strategy's payoff (fitness) \citep{weibull1995egt}. The self-organisation found in \acl{EGT} is the presence of stable steady states, in which the genotype frequencies of the population cease to change over the generations. This equilibrium is reached when all the strategies have the same expected payoff, and is called a stable steady state, because a slight perturbing will not cause a move far from the state. An \emph{\acl{ESS}} leads to a stronger asymptotically stable state, as a slight perturbing causes only a temporary move away from the state before returning \citep{weibull1995egt}. So, \acl{EGT} is focused on genetic \emph{stability} between competing between individuals, rather than the stability of the population as a whole, which therefore limits its suitability for the self-organised \emph{stability} of an evolving agent population.}

\added{\aclp{MAS} are the dominant computational technology in the evolving agent populations, and while there are several definitions of self-organisation \citep{parunak, mamei2003som, tianfield2005tso, dimarzoserugendo2006som} and stability \citep{moreau2005sms, weiss1999msm, olfatisaber2007cac} defined for \aclp{MAS}, they are not applicable primarily because of the evolutionary dynamics inherent in the context of evolving agent populations. Whereas Chli-DeWilde stability of \aclp{MAS} \citep{chli2} may be suitable, because it models \aclp{MAS} as Markov chains, which are an established modelling approach in evolutionary computing \citep{rudolph1998fmc}. A \acl{MAS} is viewed as a discrete time Markov chain with potentially unknown transition probabilities, in which the agents are modelled as Markov processes, and is considered to be \emph{stable} when its state has converged to an equilibrium distribution \citep{chli2}. Chli-DeWilde stability provides a strong notion of self-organised \emph{stability} over time, but a possible limitation is that its current formulation does not support the necessary evolutionary dynamics.}

\added{The main concept in \emph{\acl{MFT}} is that for any single particle the most important contribution to its interactions comes from its neighbouring particles \citep{parisi1998sft}. Therefore, a particle's behaviour can be approximated by relying upon the \emph{mean field} created by its neighbouring particles \citep{parisi1998sft}, and so \acl{MFT} could be suitable as a definition for the self-organised diversity of an evolving agent population. Naturally, it requires a \emph{neighbourhood model} to define interaction between neighbours \citep{parisi1998sft}, and is therefore easily applied to domains such as Cellular Automata \citep{gutowitz}. While a \emph{neighbourhood model} is feasible for biological populations \citep{flyvbjerg}, evolving agent populations lack such \emph{neighbourhood models} based on a 2D or 3D metric space, with the only available \emph{neighbourhood model} being a distance measure on a parameter space measuring dissimilarity. However, this type of \emph{neighbourhood model} cannot represent the information-based interactions between the individuals of an evolving agent population, making \acl{MFT} unsuitable as a definition for the self-organised \emph{diversity} of an evolving agent population.}

\section{\added{Complexity}}

\added{A definition for the self-organised \emph{complexity} of an evolving agent population should define the creation of coherent patterns and structures from the agents within, with no initial constraints from modelling approaches for the inclusion of pre-defined specific behaviour, but capable of representing the appearance of such behaviour should it occur.}

\added{None of the proposed definitions are directly applicable for the self-organised \emph{complexity} of an evolving agent population. The $\in$-machine modelling \citep{crutchfield2006} is not applicable, because it is only defined within the context of pre-biotic populations. Neither is the \acl{MDL} principle \citep{barron} or the Pr{\"u}gel-Bennett Shapiro formalism \citep{prugel}, because they require the involvement of \emph{subjective} human judgement at the critical stage of model and quantifier selection \citep{hansen2001msa, shapiro2001smt}. \acl{KC} complexity \citep{chaitin} is also not applicable as randomness is given maximum complexity.}

\added{Physical Complexity \citep{adami20002} fulfils abstractly the required definition for the self-organised \emph{complexity} of an evolving agent population, estimating complexity based upon the individuals of a population within the context of their environment. However, its current formulation is problematic, primarily because it is only defined for populations of fixed length, but as this is not a fundamental property of its definition \citep{adami20002} it should be feasible to redefine and extend it as needed. So, the use of Physical Complexity as a definition for the self-organised \emph{complexity} of evolving agent populations will be investigated further to determine its suitability.}

\subsection{Physical Complexity}
\label{measureSelfOrg}

\added{Understanding DNA requires knowing the environment (context) in which it exists, which may initially appear obvious as DNA is considered to be the \emph{language of life} and the purpose of life is to procreate or replicate \citep{dawkins2006sg}. Virtually all activities of biological life-forms are towards this aim \citep{dawkins2006sg}, with a few exceptions (e.g. suicide before procreation), and to achieve replication requires resources, energy and matter to be harvested. So, for any individual the environment represents the problem of extracting energy for replication, and so their DNA sequence represents a solution to this problem. Even with this understanding it would seem we still need to define the environment to be able to distinguish the information from the redundancy in a solution (DNA sequence).}

\changed{Physical Complexity was born \citep{adami1998ial} from the need to determine the proportion of information in sequences of DNA, because it has long been established that the information contained is not directly proportional to the length, known as the C-value enigma/paradox \citep{thomasjr1971goc}. However, because Physical Complexity analyses an ensemble of DNA sequences, the consistency between the different solutions shows the information, and the differences the redundancy \citep{adami2003}. Entropy, a measure of disorder, is used to determine the redundancy from the information in the ensemble. Physical Complexity therefore provides a context-relative definition for the self-organisation of a population without needing to define the context (environment) explicitly \citep{adami2000}.} \added{Furthermore, an individual DNA \emph{solution} is not necessarily a simple inverse of the \emph{problem} that the environment represents, with forms of life having evolved specialised, specific and effective ways (niches) to acquire the necessary energy and matter for replication.}

\label{defPhyCom}
Physical Complexity was derived \citep{adami2000} from the notion of \emph{conditional complexity} defined by Kolmogorov, which is different from traditional Kolmogorov complexity and states that the determination of complexity of a sequence is conditional on the environment in which the sequence is interpreted \citep{li1997ikc}. So, the complexity of a population $S$, of sequences $s$,
\begin{equation}
C = \ell - \sum\limits_{i = 1}^\ell {H(i)}, 
\label{complexity}
\end{equation}
is the maximal entropy of the population (equivalent to the length of the sequences) $\ell$, minus the sum, over the length $\ell$, of the per-site entropies $H(i)$,
\begin{equation}
H(i) = - \sum\limits_{d \in D} {p_d (i)\log _{|D|} p_d (i)}, \\
\label{persite}
\end{equation}
where $i$ is a site in the sequences ranging between one and the length of the sequences $\ell$, $D$ is the alphabet of characters found in the sequences, and $p_d(i)$ is the probability that site $i$ (in the sequences) takes on character $d$ from the alphabet $D$, with the sum of the $p_d(i)$ probabilities for each site $i$ equalling one, $\sum\limits_{d \in D} {p_d (i) = 1}$ \citep{adami2000}. \changed{So, the equivalence of the maximum complexity to the length matches the intuitive understanding that if a population of sequences of length $\ell$ has no redundancy, then their complexity is their length $\ell$.} \added{Taking the log to the base $|D|$ conveniently normalises $H(i)$ to range between zero and one.}

If $G$ represents the set of all possible genotypes constructed from an alphabet $D$ that are of length $\ell$, then the size (cardinality) of $|G|$ is equal to the size of the alphabet $|D|$ raised to the length $\ell$,
\begin{equation}
|G| = |D|^\ell.
\label{recPopSize}
\end{equation}
For the complexity measure to be accurate, a	sample size of $|D|^\ell$ is suggested to minimise the error \citep{adami2000}, but such a large quantity can be computationally infeasible. \changed{The definition's creator, for practical applications, chooses a population size of $|D|\ell $, sufficient to show any trends present.} So, for a population of sequences $S$ we choose, with the definition's creator, a computationally feasible population size of $|D|$ times $\ell $,
\begin{equation}
|S|\ \ge \ |D|\ell.
\label{popSize}
\end{equation}
The size of the alphabet, $|D|$, depends on the domain to which Physical Complexity is applied. \changed{For RNA the alphabet is the four nucleotides, $D = \{A, C, G, U\}$, and therefore $|D|=4$ \citep{adami2000}.} \added{When Physical Complexity was applied to the Avida simulation software, there was an alphabet size of twenty-eight, $|D|=28$, as that was the size of the instruction set for the self-replicating programmes \citep{adami20002}.}

\subsection{Variable Length Sequences}

Physical Complexity is currently formulated for a population of sequences of the same length \citep{adami2000}, and so we will now investigate an extension to include populations of \aclp{vls}, which will include populations of variable length agent-sequences of our Digital Ecosystem. \changed{This requires changing and re-justifying the fundamental assumptions, specifically the conditions and limits upon which Physical Complexity operates.} In (\ref{complexity}) the Physical Complexity, $C$, is defined for a population of sequences of length $\ell$ \citep{adami2000}. The most important question is what does the length $\ell$ equal if the population of sequences is of variable length? \changed{The issue is what $\ell$ represents, which is the maximum possible complexity for the population \citep{adami2000}, which will be called the \emph{complexity potential} $C_P$.} The maximum complexity in (\ref{complexity}) occurs when the per-site entropies sum to zero, $\sum\limits_{i = 1}^\ell {H(i)} \to 0$, as there is no randomness in the sites (all contain information), i.e. $C \to \ell$ \citep{adami2000}. So, the \emph{complexity potential} equals the length,
\begin{equation}
C_P = \ell,
\label{comPot2} 
\end{equation}
provided the population $S$ is of sufficient size for accurate calculations, as found in (\ref{popSize}), i.e. $|S|$ is equal or greater than $|D|\ell$. For a population of \aclp{vls}, $S_{V}$, the complexity potential, $C_{V_P}$, cannot be equivalent to the length $\ell$, because it does not exist. \changed{However, given the concept of a minimum sample size from (\ref{popSize}), there is a length for a population of \aclp{vls}, $\ell_V$, between the minimum and maximum length, such that the number of per-site samples up to and including $\ell_V$ is sufficient for the per-site entropies to be calculated.} So the \emph{complexity potential} for a population of \aclp{vls}, $C_{V_P}$, will be equivalent to its \emph{calculable} length, 
\begin{equation}
\label{potential}
C_{V_P} = \ell_V.
\end{equation}

If $\ell_V$ where to be equal to the length of the longest individual(s) $\ell _{max}$ in a population of \aclp{vls} $S_{V}$, then the operational problem is that for some of the later sites, between one and $ \ell _{max}$, the sample size will be less than the population size $|S_{V}|$. \changed{So, having the length $\ell_V$ \setCap{equal the maximum length would be incorrect}{genCap2}, as there would be an insufficient number of samples at the later sites, and therefore $\ell _V \not\equiv \ell _{max}$.} So, the length for a population of \aclp{vls}, $\ell _V $, is the highest value within the range of the minimum (one) and maximum length, $1 \le \ell _V \le \ell _{\max } $, for which there are sufficient samples to calculate the entropy. A function which provides the sample size at a given site is required to specify the value of $\ell_V$ precisely,
\begin{equation}
sampleSize(i\ :site)\ :int,
\end{equation}
where the output varies between $1$ and the population size $|S_{V}|$ (inclusive). \changed{Therefore, the length of a population of \aclp{vls}, $\ell_{V}$, is the highest value within the range of one and the maximum length, for which the sample size is greater than or equal to the alphabet size multiplied by the length $\ell _{V}$,
\begin{equation}
sampleSize(\ell _V ) \ge |D|\ell _V \wedge sampleSize(\ell _V + 1) < |D|\ell _V,
\label{lengthVLP}
\end{equation}
where $\ell _V$ is the length for a population of \aclp{vls}, and $\ell _{max} $ is the maximum length in a population of \aclp{vls}, $\ell _V$ varies between $ 1 \le \ell _V \le \ell _{max }$, $D$ is the alphabet and $|D| > 0$.} This definition intrinsically includes a minimum size for populations of \aclp{vls}, $|D|\ell _V$, and therefore is the counterpart of (\ref{popSize}), which is the minimum population size for populations of fixed length.

The length $\ell$ used in the limits of (\ref{persite}) no longer exists, and therefore (\ref{persite}) must be updated; so, the per-site entropy calculation for \aclp{vls} will be denoted by $H_{V}(i)$, and is, 
\begin{equation}
H_V (i) = - \sum\limits_{d \in D} {p_d (i)\log _{|D|} p_d (i)},
\label{perSiteVLP}
\end{equation}
where $D$ is still the alphabet, $\ell _V$ is the length for a population of \aclp{vls}, with the site $i$ now ranging between $ 1 \le i \le \ell _V $, while the $p_d (i)$ probabilities still range between $ 0 \le p_d (i) \le 1$, and still sum to one. \changed{It remains almost algebraically identical to (\ref{persite}), but the conditions and constraints of its use will change, specifically $\ell$ is replaced by $\ell_{V}$.} Naturally, $H_{V}(i)$ ranges between zero and one, as did $H(i)$ in (\ref{persite}). \changed{So, when the entropy is maximum the character found in the site $i$ is uniformly random, holding no information.}

Therefore, the complexity for a population of \aclp{vls}, $C_{V}$, is the \emph{complexity potential} of the population of \aclp{vls} minus the sum, over the length of the population of \aclp{vls}, of the per-site entropies (\ref{perSiteVLP}),
\begin{equation}
\label{newComplexity}
C_{V} = \ell _V - \sum\limits_{i = 1}^{\ell _V } {H_V (i)},
\end{equation}
where $\ell_V$ is the length for the population of \aclp{vls}, and $H_V(i)$ is the entropy for a site $i$ in the population of \aclp{vls}. 

\tfigure{width=\textwidth}{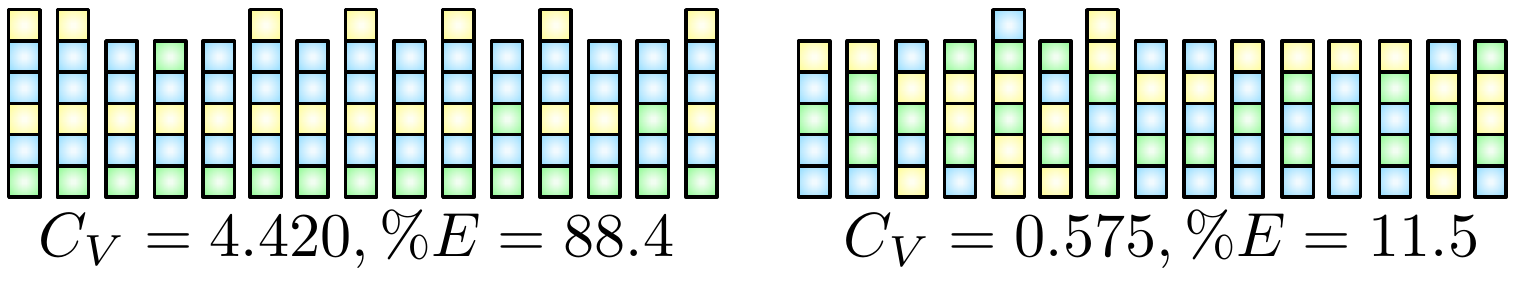}{graffle}{\changed{Abstract Visualisation for populations of Variable Length Sequences}}{\added{The Physical Complexity and Efficiency values \getCap{orgCPcap}.}}{-3mm}{!b}{}{}

\subsection{Efficiency}

\changed{Physical Complexity can now be applied to populations of \aclp{vls}, and so we will now consider the abstract example populations in Figure \ref{orgCompPop}.} We will let a single square, \squareG{-4mm}{White}{-2.5mm}, represent a site $i$ in the sequences, with different colours to represent the different values. Therefore, a sequence of sites will be represented by a sequence of coloured squares, \squareG{-4mm}{YellowGreenPurple}{-2.5mm}. \changed{Furthermore, the alphabet $D$ is the set \{\squaresub{-2mm}{Yellow}{-1.4mm}{-2pt}, \squaresub{-3.9mm}{Green}{-1.4mm}{-2pt}, \squaresub{-3.9mm}{Purple}{-1mm}{-2pt}\}, the maximum length $\ell _{max}$ is 6 and the length for populations of \aclp{vls} $\ell _V$ is calculated from (\ref{lengthVLP}) as 5.} The Physical Complexity values in Figure \ref{orgCompPop} \setCap{are consistent with the intuitive understanding one would have for the self-organisation}{orgCPcap} of the sample populations; the population with high Physical Complexity has a little randomness, while the population with low Physical Complexity is almost entirely random. 

Using our extended Physical Complexity we can construct a measure showing the use of the information space, called the Efficiency $E$, which is calculated by the Physical Complexity $C_{V}$ over the complexity potential $C_{V_P }$,
\begin{equation}
E = \frac{{C_V }}{{C_{V_P } }}.
\label{efficiencyEQ}
\end{equation}
\changed{The Efficiency $E$ will range between zero and one, reaching its maximum when the actual complexity $C_{V}$ equals the complexity potential $C_{V_{P}}$, indicating that there is no randomness in the population.} In Figure \ref{orgCompPop} the populations of sequences are shown with their respective Efficiency values as percentages, and the values are as one would expect. 

\changed{The complexity $C_{V}$ (\ref{newComplexity}) is an absolute measure, whereas the Efficiency $E$ (\ref{efficiencyEQ}) is a relative measure based on the complexity $C_{V}$.} So, the Efficiency $E$ can be used to compare the self-organised complexity of populations, independent of their size, their length, and whether their lengths are variable or not (as it is equally applicable to the fixed length populations of the original Physical Complexity). 

\subsection{\added{Clustering}}
\label{cluster123}

\added{The \emph{self-organised complexity} of an evolving agent population is the \emph{clustering}, amassing of same or similar sequences, around the optimum genome \citep{begon96}. This can be visualised on a \emph{fitness landscape} \citep{wright1932}, which \setCap{shows the combination space (power set) of the alphabet $D$ against the fitness values from the \emph{selection pressure} (user request)}{FLsingleCap}. \setCap{The agent-sequences of an evolving population will evolve, clustering around the optimal genome}{FLsingle2Cap}, assuming that its evolutionary process does not become trapped while clustering over local optima, and as shown in Figure \ref{fitLandSingleConverted}.}

\tfigure{width=3.5in}{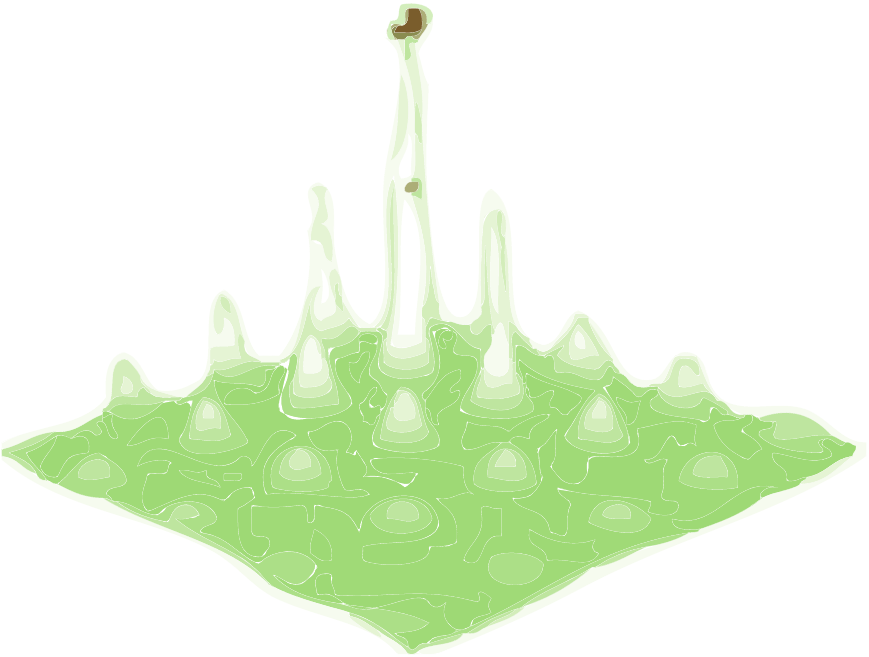}{ai}{\added{3D Fitness Landscape with a Global Optimum}}{\added{This \getCap{FLsingleCap}, resulting in a global optimum.}}{0mm}{!h}{}{}

\added{Clustering is indicated by the Efficiency $E$ tending to its maximum, as the population's Physical Complexity $C_{V}$ tends to the \emph{complexity potential} $C_{V_{P}}$, because an optimal sequence is becoming dominant in the population, and therefore increasing the uniformity of the sites across the population. With a global optimum, the Efficiency $E$ tends to a maximum of one, indicating that the \emph{evolving population of sequences} is tending to a \emph{set of clusters $T$ of size one},
\begin{equation} 
E = \frac{C_V}{C_{V_P}} \to 1 \ as\ |T| \to 1,
\label{clusters1}
\end{equation}
assuming its evolutionary process does not become trapped at local optima. So, the \emph{tending} of the Efficiency $E$ provides a \emph{clustering coefficient}. It \emph{tends}, never quite reaching its maximum, because of the mutation inherent in the evolutionary process.}

\added{The other extreme scenario occurs when the number of clusters equals the size of the population, which would only occur with a \emph{flat fitness landscape} \citep{kimura:ntm} resulting from a non-discriminating \emph{selection pressure}, as shown in Figure \ref{fitLandFlatConverted}. The population occupancy is \setCap{uniformly random}{FLflatCap}, as any position (sequence) has the same fitness as any other. So the entropy (randomness) tends to maximum, resulting in the complexity $C_{V}$ tending to zero, and therefore the Efficiency $E$ also tending to zero, while the number of clusters $|T|$ tends to the number of sequences in the population $|S|$,
\begin{eqnarray}
E = \frac{{C_V }}{{C_{V_P } }} \to 0 \ as \ |T| \to |S|.
\label{clusters0}
\label{maxClusters}
\end{eqnarray}
So the number of clusters $|T|$ tends to the population size $|S|$, with each cluster consisting of only one unique sequence (individual).}

\tfigure{width=3.5in}{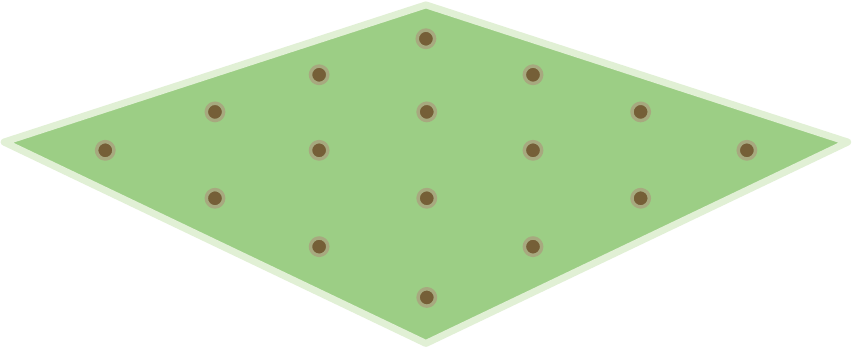}{graffle}{\added{3D Fitness Landscape with No Optimum}}{\added{Theoretical extreme scenario in which the selection pressure is non-discriminating. So, the population occupancy of the fitness landscape would then be \getCap{FLflatCap}.}}{0mm}{}{}{}

\tfigure{width=3.5in}{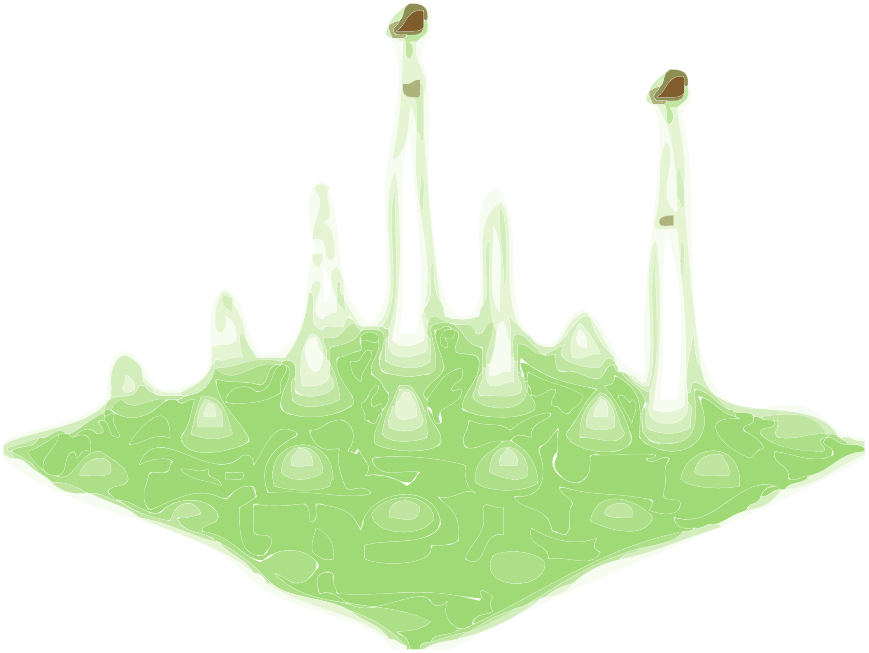}{ai}{\added{3D Fitness Landscape with Global Optima}}{\added{Clustering scenario, in which the Efficiency $E$ of the population $S$ tends to a value based on the number of clusters $|T|$, because of clustering around multiple optima.}}{0mm}{!b}{}{}

\added{If there are global optima, as there are in Figure \ref{fitLandMultipleConverted}, the \setCap{Efficiency $E$ will tend to a maximum below one, because the population of sequences consists of more than one cluster, with each having an Efficiency tending to a maximum of one.}{FLmul2Cap} \setCap{The simplest scenario of clusters is \emph{pure clusters}}{FLmulCap}; \emph{pure} meaning that each cluster uses a distinct (mutually exclusive) subset of the alphabet $D$ relative to any other cluster. In this scenario the Efficiency $E$ tends to a value based on the number of clusters $|T|$, because a \emph{number} of the $p_d (i)$ probabilities at each \emph{site} in (\ref{perSiteVLP}) are the reciprocal of the number of clusters, $\frac{1}{|T|}$. So, given that the \emph{number} of the $p_d (i)$ probabilities taking the value $\frac{1}{|T|}$ is equal to the number of clusters, while the other $p_d (i)$ probabilities take a value of zero, then the per-site entropy calculation of $H_V (i)$ from (\ref{perSiteVLP}) becomes
\begin{equation}
H_V (i) = \log _{|D|} |T|,
\label{calcNumClusters}
\end{equation}
where $i$ is the site, $|D|$ is the alphabet size, and $|T|$ is the number of clusters. Hence, given (\ref{calcNumClusters}), (\ref{newComplexity}), and (\ref{potential}), then the Efficiency $E$ from (\ref{efficiencyEQ}) becomes 
\begin{equation}
E \to 1 - (\log _{|D|} |T|),
\label{calcNumClusters2}
\end{equation}
where $|D|$ is the alphabet size and $|T|$ is the number of clusters. Therefore, the Efficiency $E$, the \emph{clustering coefficient}, tends to a value that can be used to determine the number of \emph{pure clusters} in an evolving population of sequences.}

\added{For a population $S$ with clusters, each cluster is a sub-population with an Efficiency $E$ tending to a maximum of one. To specify this relationship we require a function that provides the Efficiency $E$ (\ref{efficiencyEQ}) of a population or sub-population of sequences,
\begin{equation}
\mbox{\emph{efficiency(input :population) :int}}.
\end{equation}
So, for a population $S$ consisting of a set of clusters $T$, each member (cluster) $t$ is therefore a sub-population of the population $S$, and is defined as
\begin{eqnarray}
&t \in T \to & \\ \nonumber
&\left(t \subseteq S \wedge \mbox{\textit{efficiency(t)}} \to 1 \wedge |t| \approx \frac{|S|}{|T|} \wedge \sum\limits_{t \in T} {|t|} = |S|\right),&
\label{defineCluster}
\end{eqnarray}
where a cluster $t$ has an Efficiency $E$ tending to a maximum of one, and the cluster size $|t|$ is approximately equal to the population size $|S|$ divided by the number of clusters $|T|$. It is only \emph{approximately equal} because of variation from mutation, and because the population size may not divide to a whole number. These conditions are true for all members $t$ of the set of clusters $T$, and therefore the summation of the cluster sizes $|t|$ equals the size of the population $|S|$.}

\tfigure{scale=0.8}{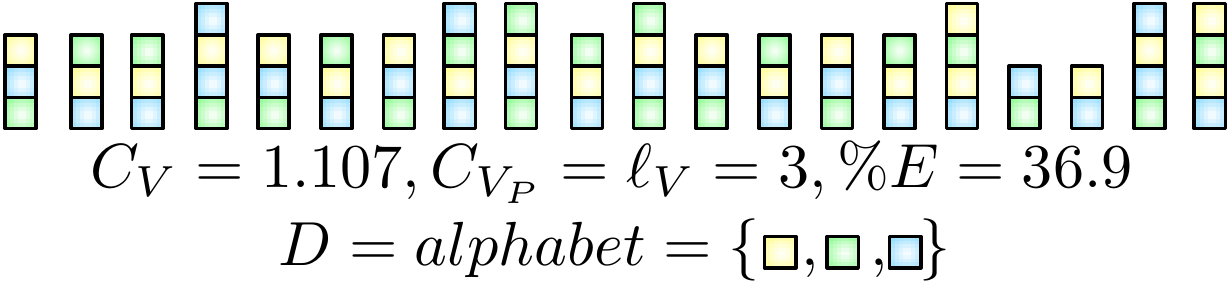}{graffle}{\added{Population with Hidden Clusters}}{\added{Visualisation for the population of sequences from a population with global optima, with clusters visually hard to identify.}}{0mm}{!b}{}{}

\tfigure{scale=0.8}{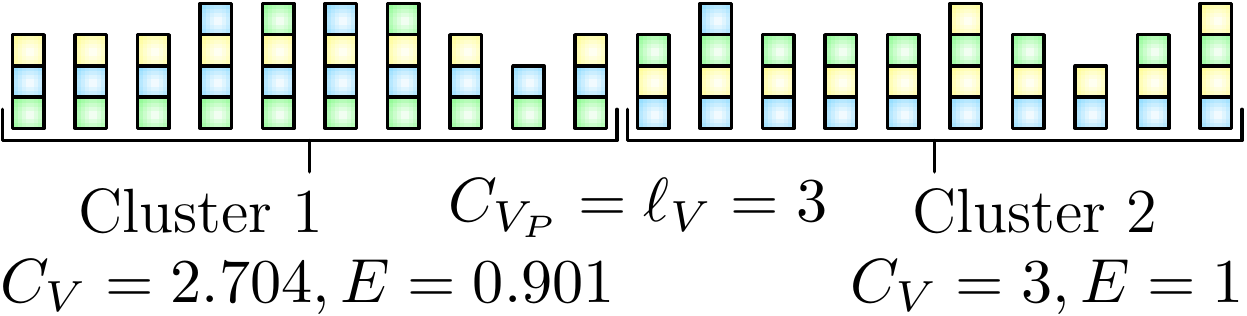}{graffle}{\added{Population with Clusters Visible}}{\added{Visualisation for a population of sequences with global optima, which has been arranged to show the clusters present.}}{0mm}{!t}{}{}

\added{The population of sequences from the \emph{fitness landscape} of Figure \ref{fitLandMultipleConverted} is visualised in Figure \ref{populationHiddenClusters}, but the clusters within cannot be seen. So, the population is arranged to show the clustering in Figure \ref{populationWithClusters}, in which the two clusters are clearly evident. \setCap{The clusters of the population have Efficiency values tending to a maximum of one, compared to the Efficiency of the population as a whole, which is tending to a maximum significantly below one.}{popClusShowCap} This is the expected behaviour of clusters as defined in (\ref{defineCluster}).}

\added{The population size $|S|$, in Figure \ref{populationWithClusters}, is double the minimum requirement specified in (\ref{lengthVLP}), so that the complexity $C_{V}$ (\ref{newComplexity}) and Efficiency $E$ (\ref{efficiencyEQ}) could be used in defining the principles of clustering without redefining the \emph{length of a population of \aclp{vls}} $\ell_{V}$ (\ref{lengthVLP}). However, when determining the variable length $\ell_{V}$ of a cluster $t$, the sample size requirement is different, specifically a cluster $t$ is a sub-population of $S$, and therefore by definition cannot have a population size equivalent to $S$ (unless the population consists of only one cluster). Therefore, to manage clusters requires a reformulation of $\ell _V$ (\ref{lengthVLP}) to
\begin{equation}
\ell _V = \left(
\begin{array}{l}
sampleSize(\ell _V ) \approx \frac{|D|\ell _V }{|T|} \wedge \\
\vspace{-3mm}\white{.} \\
sampleSize(\ell_{V} + 1) < \frac{|D|\ell _V }{|T|} \\
\end{array}\right),
\label{clustersSampleSize}
\end{equation}
where $\ell _{max} $ is the maximum length in a population of \aclp{vls}, $\ell _V$ varies between $ 1 \le \ell _V \le \ell _{max }$, $D$ is the alphabet, $|D| > 0$, and $T$ is the set of clusters in the population $S$.}

\added{A population with clusters will always have an Efficiency $E$ tending towards a maximum significantly below one. Therefore, managing populations with clusters requires a reformulation of the Efficiency (\ref{efficiencyEQ}) to}
\added{\begin{equation}
E_{c} (S) = \left\{
\begin{array}{cl}
\frac{C_V }{C_{V_P}} & \mbox{if $|T| = 1$} \\
\vspace{-2mm}\white{.} & \white{.}\\
\frac{\sum\limits_{t \in T} {E_{c} (t)}}{|T|} & \mbox{if $|T| > 1$}
\end{array}\right.,
\label{efficiencyMultiple}
\end{equation}}
\added{where $t$ is a cluster, and a member of the set of clusters $T$ of the population $S$. So, the Efficiency $E_{c}$ is equivalent to the Efficiency $E$ if the population consists of only one cluster, but if there are clusters then the Efficiency $E_{c}$ is the average of the Efficiency $E$ values of the clusters.}

\subsection{\added{Atomicity}}
\label{atomicitySection}

\added{\emph{Atomicity} is the property \setCap{of a set of agents, such that no single agent can functionally replace any agent-sequence, i.e. their functionality is \emph{mutually exclusive} to one another.}{atomCap} It is important because non-atomicity can adversely affect the uniformity of the calculated per-site entropies, which is the main construct of \setCap{the Physical Complexity measure}{atom2Cap}, and so non-atomicity risks introducing error when calculating the information content. Our extensions to Physical Complexity to support clustering are also necessary to manage non-atomicity, because it leads to the formation of clusters within evolving agent populations. The presence of clusters can be identified by the \emph{clustering coefficient}, the Efficiency $E$ tending to a value below one, with the Efficiency $E_{c}$ (\ref{efficiencyMultiple}) used to calculate the actual Efficiency as it supports clustering and therefore non-atomicity.}

\added{If we consider the example population shown in Figure \ref{nonAtomicPopulation}, which is constructed from an alphabet in which the yellow agent \squareG{-3mm}{Yellow}{-2mm} can functionally replace a green blue agent-sequence \squareG{-3.5mm}{GreenBlue}{-2.5mm}, and so the uniformity across site two is lost. Therefore, the \setCap{Efficiency $E$ of the population is a half, whereas the Efficiency $E_{c}$ for populations with clusters is one, because it supports clustering and therefore non-atomicity}{popAtomCap}.}

\tfigure{scale=0.9}{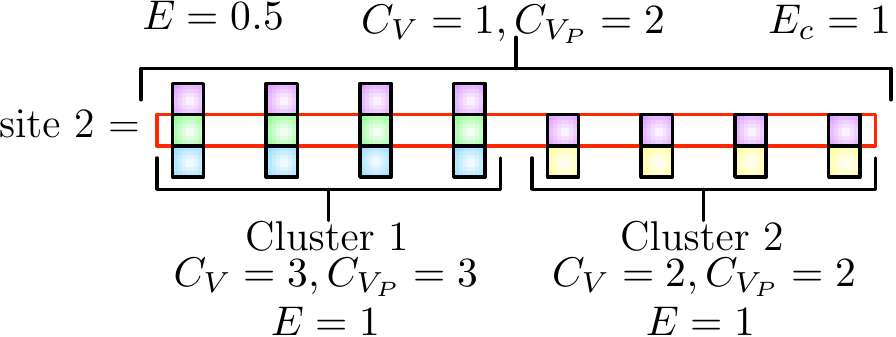}{graffle}{\added{Population Constructed from a Non-Atomic Alphabet}}{\added{The population is constructed from an alphabet in which the yellow agent is able to functionally replace a green blue agent-sequence.}}{0mm}{!h}{}{-8mm}

\section{\added{Stability}}

\added{A definition for the self-organised \emph{stability} of an evolving agent population should define the resulting stability or instability that emerges over time, with no initial constraints from modelling approaches for the inclusion of pre-defined specific behaviour, but capable of representing the appearance of such behaviour should it occur.}

\added{None of the proposed definitions are directly applicable for the self-organised \emph{stability} of an evolving agent population. The $\in$-machine modelling \citep{crutchfield2006} is not applicable, because it is only defined within the context of pre-biotic populations. The Pr{\"u}gel-Bennett Shapiro formalism \citep{prugel} is not suitable, because it necessitates the involvement of \emph{subjective} human judgement at the critical stage of quantifier selection. \acl{SOC} \citep{bak1988soc} is also not applicable as it only models the events of genetic change in the population over time, rather than measuring the resulting stability or instability of the population. Neither is \acl{EGT} \citep{weibull1995egt}, which only defines the \emph{genetic stability} of the genotypes, in terms of \emph{equilibrium} and \emph{non-equilibrium} dynamics, instead of the stability of the population as a whole.}

\added{Chli-DeWilde stability of \aclp{MAS} \citep{chli2} does fulfil the required definition of the self-organised \emph{stability}, measuring convergence to an equilibrium distribution. However, its current formulation does not include \aclp{MAS} that make use of \emph{evolutionary computing} algorithms, i.e. our evolving agent populations, but it could be extended to include such \aclp{MAS}, because its \emph{Markov-based modelling} approach is well established in \emph{evolutionary computing} \citep{rudolph1998fmc}. While there has been past work on modelling \emph{evolutionary computing} algorithms as Markov chains \citep{rudolph, nix, goldberg2, eibenAarts}, we have found none including \aclp{MAS} despite both being mature research areas \citep{masOverviewPaper, ecpaper}, because their integration is a recent development \citep{smith1998fec}. So, the use of Chli-DeWilde stability as a definition for the self-organised \emph{stability} of evolving agent populations will be investigated further to determine its suitability.}

\subsection{\changed{Chli-DeWilde Agent Stability}}

\added{We will now briefly introduce Chli-DeWilde stability for \acl{MAS} and Evolutionary Computing, sufficiently to allow for the derivation of our extensions to Chli-DeWilde stability to include \aclp{MAS} with Evolutionary Computing.} Chil-DeWilde stability was created to provide a clear notion of stability in \acp{MAS} \citep{chli2}, because stability is perhaps one of the most desirable features of any engineered system, given the importance of being able to predict its response to various environmental conditions prior to actual deployment; and while computer scientists often talk about stable or unstable systems \citep{mspaper5ThomasSycara1998, mspaper9Balakrishnan1997}, they did so without having a concrete or uniform definition of stability. \changed{Also, other properties had been widely investigated, such as openness \citep{mspaperAbramov2001}, scalability \citep{mspaperMarwala2001} and adaptability \citep{mspaperSimoesMarques2003}, but not stability.} So, the Chli-DeWilde definition of stability for \acp{MAS} was created \citep{chli2}, based on the stationary distribution of a stochastic system, modelling the agents as Markov processes, and therefore viewing a \ac{MAS} as a discrete time Markov chain with a potentially unknown transition probability distribution. \changed{The \ac{MAS} is stable once its state, a stochastic process, has converged to an equilibrium distribution \citep{chli2}, because stability of a system can be understood intuitively as exhibiting bounded behaviour.}

Chli-DeWilde stability was derived \citep{chlithesis} from the notion of stability defined by De Wilde \citep{mspaperDeWilde1999a, mspaperLee1998}, based on the stationary distribution of a stochastic system, making use of discrete-time Markov chains, which we will now introduce\added{\arxivfootnote{\added{A more comprehensive introduction to Markov chain theory and stochastic processes is available in \citep{msthesisNorris1997} and \citep{msthesisCoxMiller1972}.}}}. \changed{If we let $I$ be a \emph{countable set}, such that each $i \in I$ is called a \emph{state} and $I$ is called the \emph{state-space}.} We can then say that $\lambda = (\lambda_i : i \in I)$ is a \emph{measure on} $I$ if $0 \le \lambda_i < \infty$ for all $i \in I$, and additionally a \emph{distribution} if $\sum_{i \in I}{\lambda_i=1}$ \citep{chlithesis}. \changed{So, if $X$ is a \emph{random variable} taking values in $I$ and we have $\lambda_i = \Pr(X = i)$, then $\lambda$ is \emph{the distribution of $X$}, we can say that a matrix $P = (p_{ij} : i,j \in I)$ is \emph{stochastic} if every row $(p_{ij} : j \in I)$ is a \emph{distribution} \citep{chlithesis}.} We can then extend familiar notions of matrix and vector multiplication to cover a general index set $I$ of potentially infinite size, by defining the multiplication of a matrix by a measure as $\lambda P$, which is given by
\begin{equation}
(\lambda P)_i = \sum\limits_{j \in I}{\lambda_{j}p_{ij}}.
\label{ms3dot1}
\end{equation}
We can now describe the rules for a Markov chain by a definition in terms of the corresponding matrix $P$ \citep{chlithesis}.\\

\begin{definition}
We say that $(X^t)_{t\ge0}$ is a Markov chain with initial distribution $\lambda = (\lambda_i : i \in I)$ and transition matrix $P = (p_{ij} : i,j \in I)$ if:
\narrowlinespacing
\begin{enumerate}
\item $\Pr(X^0 = i_0) = \lambda_{i_0}$ and
\item $\Pr(X^{t+1} = i_{t+1}\ |\ X^0 = i_0, \ldots, X^t = i_t) = p_{i_t i_{t+1}}$.
\end{enumerate}
\vspace{-3mm}
\normallinespacing
We abbreviate these two conditions by saying that $(X^t)_{t\ge0}$ is Markov$(\lambda, P)$.
\end{definition}

In this first definition the Markov process is \emph{memoryless}\added{\footnote{\added{Markov systems with probabilities is a very powerful modelling technique, applicable in large variety of scenarios, and it is common to start memoryless, in which the output probability distribution only depends on the current input. However, there are scenarios in which alternative modelling techniques, like queueing systems, are more suitable, such as when there is asynchronous communications, and to fully characterise the system state at time (t), the history of states at (t-1), (t-2), ... might also need to be considered.}}}, resulting in only the current state of the system being required to describe its subsequent behaviour. \changed{So, we say that a Markov process $X^0, X^1, \ldots, X^t$ has a \emph{stationary distribution} if the probability distribution of $X^t$ becomes independent of the time $t$ \citep{chli2}. Therefore}, the following theorem is an \emph{easy consequence} of the second condition from the first definition.\\

\begin{theorem}
A discrete-time random process $(X^t)_{t\ge0}$ is Markov$(\lambda,P)$, if and only if for all $t$ and $i_0, \ldots, i_t$ we have
\narrowlinespacing
\begin{equation}
\Pr(X^0 = i_0, \ldots, X^t = i_t) = \lambda_{i_0}p_{i_0 i_1} \cdots p_{i_{t-1}i_t}.
\label{ms3dot2}
\end{equation}
\vspace{-6mm}
\normallinespacing
\end{theorem}
 
This first theorem depicts the structure of a Markov chain, \added{\citep{chlithesis,msthesisNorris1997,msthesisCoxMiller1972}, illustrating the relation with the stochastic matrix $P$. The next Theorem shows how the Markov chain evolves in time, again showing the role of the matrix $P$.}\\

\begin{theorem}
Let $(X^t)_{t\ge0}$ be $Markov(\lambda,P)$, then for all $t,s\ge0$:
\narrowlinespacing
\begin{enumerate}
\item $\Pr(X^t = j) = (\lambda P^t)_j$ and
\item $\Pr(X^t = j\ |\ X^0 = i) = \Pr(X^{t+s} = j\ |\ X^s = i) = (P^t)_{ij}$.
\end{enumerate}
\vspace{-3mm}
\normallinespacing
\label{ms3dot3dot2}
For convenience $(P^t)_{ij}$ can be more conveniently denoted as $p^{(t)}_{ij}$.
\end{theorem}

\changed{Given this second theorem we can define $p^{(t)}_{ij}$ as the t-step transition probability from the state $i$ to $j$ \citep{chlithesis}, so we can now introduce the concept of an \emph{invariant distribution} \citep{chlithesis}, in which we say that $\lambda$ is invariant if
\begin{equation}
\lambda P = \lambda .
\end{equation}}
The next theorem will link the existence of an \emph{invariant distribution}, which is an algebraic property of the matrix $P$, with the probabilistic concept of an \emph{equilibrium distribution}. This only applies to a restricted class of Markov chains, namely those with \emph{irreducible} and \emph{aperiodic} stochastic matrices. \changed{However, there is a multitude of analogous results for other types of Markov chains to which we can refer \citep{msthesisNorris1997, msthesisCoxMiller1972}, and the following theorem is provided as an exmaple of the family of theorems that apply.} An \emph{irreducible} matrix $P$ is one for which, for all $i,j \in I$ there are sufficiently large $t,p^{(t)}_{ij} > 0$, and is \emph{aperiodic} if for all states $i \in I$ we have $p^{(t)}_{ii} > 0$ for all sufficiently large $t$ \added{\citep{chlithesis,msthesisNorris1997,msthesisCoxMiller1972}}. \added{The meaning of these properties can broadly be explained as follows. An irreducible Markov chain is a chain where all states intercommunicate. For this to happen, there needs to be a non-zero probability to go from any state to any other state. This communication can happen in any number $t$ of time steps. This leads to the condition $p^{(t)}_{ij} > 0$ for all $i$ and $j$. An aperiodic Markov chain is a chain where all states are aperiodic. A state is aperiodic if it is not periodic. Finally, a state is periodic if subsequent occupations of this state occur at regular multiples of a time interval. For this to happen, $p^{(t)}_{ii}$ has to be zero for $t$ an integral multiple of a number. This leads to the condition $p^{(t)}_{ii} > 0$ for {\em a}-periodicity. For further explanations, please refer to \citep{msthesisCoxMiller1972}.}\\

\begin{theorem}
Let $P$ be irreducible, aperiodic and have an invariant distribution, $\lambda$ can be any distribution, and suppose that $(X^t)_{t\ge0}$ is Markov$(\lambda, P)$ \citep{chlithesis}, then
\narrowlinespacing
\begin{eqnarray}
& \Pr(X^t = j) \to p_{j}^\infty \ as\ t \to \infty\ \mbox{for all}\ j \in I & \\
& and & \nonumber \\
& p^{(t)}_{ij} \to p_{j}^\infty \ as\ t \to \infty\ \mbox{for all}\ i,j \in I. &
\end{eqnarray}
\vspace{-9mm}
\normallinespacing
\end{theorem} 
 
\changed{We can now view a system $S$ as a countable set of states $I$ with implicitly defined transitions $P$ between them, such that at time $t$ the state of the system is the random variable $X^t$, with the key assumption that $(X^t)_{t,0}$ is Markov$(\lambda,P)$ \added{\citep{chlithesis,msthesisNorris1997,msthesisCoxMiller1972}}.}\\

\begin{definition}
The system $S$ is said to be stable when the distribution of the its states converge to an \emph{equilibrium distribution},
\narrowlinespacing
\begin{equation}
\Pr(X^t = j) \to p_{j}^\infty \ as\ t \to \infty\ for\ all j\ \in I.
\end{equation}
\vspace{-9mm}
\normallinespacing
\end{definition}

More intuitively, the system $S$, a stochastic process $X^0$,$X^1$,$X^2$,... is \emph{stable} if the probability distribution of $X^t$ becomes independent of the time index $t$ for large $t$ \citep{chli2}. \changed{Most Markov chains with a finite state-space and positive transition probabilities are examples of stable systems, because after an initialisation period they stabalise on a stationary distribution \citep{chlithesis}.}

A \ac{MAS} can be viewed as a system $S$, with the system state represented by a finite vector $\bx$, having dimensions large enough to manage the agents present in the system. The state vector will consist of one or more elements for each agent, and a number of elements to define general properties\added{\footnote{\added{These general properties are intended to represent properties that are external to the agents, and as such could include the coupling between the agents. However, we would expect such properties, as the coupling between the agents, to be stored within the agents themselves, and so be part of the elements defining the agents.}}} of the system state. \added{Hence there are many more states of the system (different state vectors) than there are agents.} 

\subsection{Extensions for Evolving Populations}
\label{def}

\added{Having now introduced Chli-DeWilde stability, we will now consider the Evolutionary Computing of our Digital Ecosystems in greater detail, sufficiently to allow for the derivation of our extensions to Chli-DeWilde stability to include our Digital Ecosystems.}

\added{This Evolutionary Computing is now recognised as a sub-field of artificial intelligence (more particularly computational intelligence) that involves combinatorial optimisation problems \citep{ec17}. Evolutionary algorithms are based upon several fundamental principles from biological evolution, including reproduction, mutation, recombination (crossover), natural selection, and survival of the fittest. As in biological populations, evolution occurs by the repeated application of the above operators \citep{back1996eat}. An evolutionary algorithm operates on the collection of individuals making up a population. An \emph{individual}, in the natural world, is an organism with an associated fitness \citep{lawrence1989hsd}. So, candidate solutions to an optimisation problem play the role of individuals in a population, and a cost (fitness) function determines the environment within which the solutions \emph{live}, analogous to the way the environment selects for the fittest individuals. The number of individuals varies between different implementations and may also vary during the use of an evolutionary algorithm. Each individual possesses some characteristics that are defined through its genotype, its genetic composition, which will be passed onto the descendants of that individual \citep{back1996eat}. Processes of mutation (small random changes) and crossover (generation of a new genotype by the combination of components from two individuals) may occur, resulting in new individuals with genotypes differing from the ancestors they will come to replace. These processes iterate, modifying the characteristics of the population \citep{back1996eat}. Which members of the population are kept, and used as parents for offspring, depends on the fitness (cost) function of the population. This enables improvement to occur \citep{back1996eat}, and corresponds to the fitness of an organism in the natural world \citep{lawrence1989hsd}. Recombination and mutation create the necessary diversity and thereby facilitate novelty, while selection acts as a force increasing quality. Changed pieces of information resulting from recombination and mutation are randomly chosen. Selection operators can be either deterministic, or stochastic. In the latter case, individuals with a higher fitness have a higher chance to be selected than individuals with a lower fitness \citep{back1996eat}.}

\changed{So, extending Chli-DeWilde stability to the \emph{class} of \acp{MAS} that make use of \emph{evolutionary computing} algorithms, including our evolving agent populations, requires consideration of the following issues: the inclusion of \emph{population dynamics}, and an understanding of population \emph{macro-states}.}

\subsubsection{Population Dynamics}

First, the \ac{MAS} of an evolving agent population is composed of $n$ agent-sequences, with each agent-sequence $i$ in a state $\xi_i^t$ at time $t$, where $i=1, 2, \ldots, n$. \changed{The states of the agent-sequences are \emph{random variables}, so that the state vector for the \ac{MAS} is a vector of random variables $\bxi^t$, with the time being discrete, $t=0, 1, \ldots$ .} The interactions among the agent-sequences are noisy, and are given by the probability distributions
\be
\Pr(X_i | \by) = \Pr(\xi_i^{t+1} = X_i | \bxi^t = \by) , \quad \ra,
\eeq{eq1}
where $X_i$ is a value for the state of agent-sequence $i$, and $\by$ is a value for the state vector of the \ac{MAS}. The probabilities implement a Markov process, with the noise caused by mutations. \changed{Furthermore, the agent-sequences are each subjected to the \emph{selection pressure} from the environment of the system, which is applied equally to all the agent-sequences of the population.} So, the probability distributions are statistically independent, and
\be
\Pr(\bx | \by) = \Pi_{i=1}^n \Pr(\xi_i^{t+1} = X_i | \bxi^t = \by).
\eeq{eq5}
If the occupation probability of state $\bx$ at time $t$ is denoted by $p_{\bx}^t$, then
\be
p_{\bx}^t = \sum_{\by} \Pr(\bx | \by) p_{\by}^{t-1}.
\eeq{eq5.1}
This is a discrete time equation used to calculate the evolution of the state occupation probabilities from $t=0$, while equation (\ref{eq5}) is the probability of moving from one state to another. \changed{The \ac{MAS} (evolving agent population) is self-stabilising if the limit distribution, of the occupation probabilities, exists and is non-uniform}, i.e.
\be
p_\bx^\infty = lim_{t \rightarrow \infty} p_{\bx}^t
\eeq{eq2}
exists for all states $\bx$, and there exist states $\bx$ and $\by$ such that
\be
p_\bx^\infty \neq p_\by^\infty.
\eeq{eq3}
These equations define that some configurations of the system, after an extended time, will be more likely than others, because the likelihood of their occurrence no longer changes. \changed{Such a system is \emph{stable}, because the occurrence of states no longer changes with time, and is the definition of stability developed in \citep{chli2}.} While equation (\ref{eq2}) is the \emph{probabilistic equivalence} of an \emph{attractor}\footnote{An attractor is a set of states, invariant under the dynamics, towards which neighbouring states asymptotically approach during evolution.} in a system with deterministic interactions, which we had to extend to a stochastic process because mutation is inherent in evolutionary dynamics.

\changed{While the number of agents in the Chli-DeWilde formalism varies, we require variation according to the \emph{selection pressure} acting upon the evolving agent population.} We must therefore formally define and extend the definition of \emph{dead} agents, by introducing a new state $d$ for each agent-sequence. If an agent-sequence is in this state, $\xi_i^t=d$, then it is \emph{dead} and does not affect the state of other agent-sequences in the population. If an agent-sequence $i$ has low fitness then that agent-sequence will likely die, because
\be
\Pr(d | \by) = \Pr(\xi_i^{t+1} = d | \bxi^t = \by)
\eeq{eq4}
will be high for all $\by$. \changed{Conversely, if an agent-sequence has high fitness, it will likely replicate and assume the state of a similarly successful agent-sequence (mutant), or crossover might occur changing the state of the successful agent-sequence and another agent-sequence.}

\subsubsection{Population Macro-States}

\changed{The state of the system, an evolving agent population, $S$ is determined by the collection of agents of which it consists at a specific time $t$, which potentially changes as the time increases, $t+1$.} \added{This collection of agents will have varying fitness values, and so the one with the highest fitness at the current time $t$ is the \emph{current maximum fitness individual}. For example, an evolving agent population with individuals ranging in fitness between 36.2\% and and 45.8\%, the \emph{current maximum fitness individual (agent)} is the one with a fitness of 45.8\%.} \changed{So, we can define a macro-state $M$ as a set of states (evolving agent populations) with a common property, here possessing at least one copy of the \emph{current maximum fitness individual}. Therefore, by its definition, each macro-state $M$ must also have a \emph{maximal state} composed entirely of copies of the \emph{current maximum fitness individual}.} \added{If the population size is not fixed (not in nature, can be in evolutionary computing), the state space of the evolving agent population is infinite, but in practise would be bounded by resource availability. So, there is also an infinite number of configurations for an evolving agent population that has the same \emph{current maximum fitness individual}.}

\added{So, the state-space $I$ of the system (evolving agent population) $S$ can be grouped to a set macro-states $\{M\}$.} \changed{For one macro-state, which we will call the \emph{maximum macro-state} $M_{max}$, the \emph{current maximum fitness individual} will be the \emph{global maximum fitness individual}, which is the \emph{optimal solution} (\emph{fittest} individual) that the evolutionary computing process can reach through the evolving agent population (system) $S$.} \added{For example, an evolving agent population at its \emph{maximum macro-state} $M_{max}$, with individuals ranging in fitness between 88.8\% and and 96.8\%, the \emph{global maximum fitness individual (agent)} is the one with a fitness of 96.8\% and there will be no \emph{fitter} agent. Also, we can therefore refer to all other macro-states of the system $S$ as \emph{sub-optimal} macro-states, as there can be only one \emph{maximum macro-state} $M_{max}$.}

\tfigure{scale=0.8}{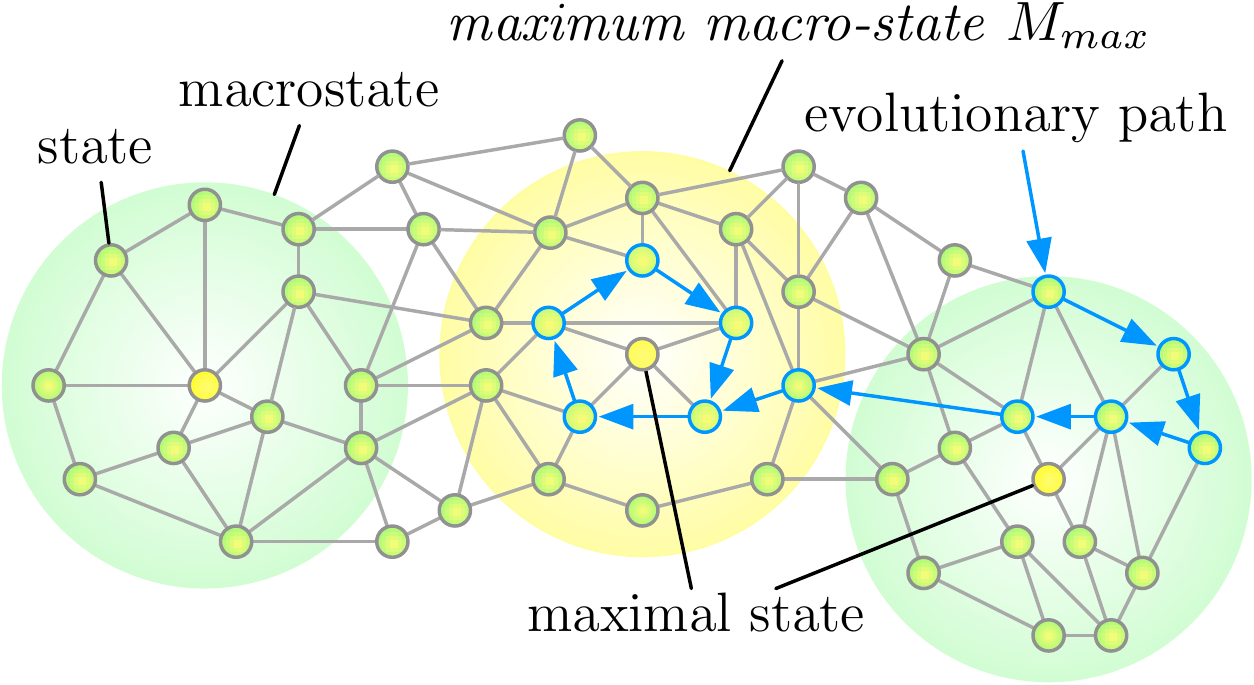}{graffle}{State-Space of an Evolving Agent Population}{A \getCap{statesCap} is shown, with \getCap{capStates3} \getCap{capStates}.}{0mm}{!b}{}{}

We can consider the \emph{macro-states} of an evolving agent population visually through the representation of the state-space $I$ of the system $S$ shown in Figure \ref{states}, which includes a \setCap{possible evolutionary path through the state-space $I$}{statesCap}. \changed{Traversal through the state-space $I$ is directed by \setCap{the \emph{selection pressure} of the evolutionary process}{capStates3} acting upon the population $S$, \setCap{driving it towards the \emph{maximal state} of the \emph{maximum macro-state} $M_{max}$}{capStates}, which consists entirely of copies of the \emph{optimal solution}, and is the equilibrium state that the system $S$ is forever \emph{falling towards} without ever quite reaching, because of the mutation (noise) within the system.} So, while this \emph{maximal state} will never be reached, the \emph{maximum macro-state} $M_{max}$ itself is certain to be reached, provided the system does not get trapped at local optima, i.e. the probability of being in the \emph{maximum macro-state} $M_{max}$ at infinite time is one, $p^{\infty}_{\bone}=1$, as defined from equation (\ref{eq5.1}).

\changed{Furthermore, we can define quantitatively the probability distribution of the macro-states that the system will occupy at infinite time.} For a stable system, as defined by equation (\ref{eq3}), the \emph{degree of instability}, $d_{ins}$, can be defined as the entropy of its probability distribution at infinite time,
\be
d_{ins} = H(p^\infty) = -\sum\limits_{\bx}p_{\bx}^{\infty}log_{N}(p_{\bx}^{\infty}),
\eeq{eq6}
where $N$ is the number of possible states, and taking $log$ to the base $N$ normalises the \emph{degree of instability}. \changed{The \emph{degree of instability} will range from zero (inclusive) and one (exclusive), because the maximum instability of one would only occur during the theoretical extreme scenario of a \emph{non-discriminating selection pressure}.}

\section{Diversity}

\added{A definition for the self-organised \emph{diversity} of an evolving agent population should define the optimal variability, of the agents and agent-sequences, that emerge over time, with no initial constraints from modelling approaches for the inclusion of pre-defined specific behaviour, but capable of representing the appearance of such behaviour should it occur.}

\added{None of the proposed definitions are applicable for the self-organised \emph{diversity} of an evolving agent population. The $\in$-machine modelling \citep{crutchfield2006} is not applicable, because it is only defined within the context of pre-biotic populations. Neither is the \acl{MDL} \citep{barron} principle or the Pr{\"u}gel-Bennett Shapiro formalism \citep{prugel} suitable, because they necessitate the involvement of \emph{subjective} human judgement at the critical stages of model or quantifier selection. \acl{MFT} is also not applicable because of the necessity of a \emph{neighbourhood model} for defining interaction, and evolving agent populations lack a 2D or 3D metric space for such models. So, the only available \emph{neighbourhood model} becomes a distance measure on a parameter space that measures dissimilarity. However, this type of \emph{neighbourhood model} cannot represent the information-based interactions between the individuals of an evolving agent population.}

\added{We suggest that the uniqueness of Digital Ecosystems makes the application of existing definitions inappropriate for the self-organised \emph{diversity}, because while we could extend a biology-centric definition for the self-organised \emph{complexity}, and a computing-centric definition for the self-organised \emph{stability}, we found neither of these approaches, or any other, appropriate for the self-organised \emph{diversity}. The Digital Ecosystem being the digital counterpart of a biological ecosystem gives it unique properties. So, the evolving agent populations possess properties of both computing systems (e.g. agent systems) as well as biological systems (e.g. population dynamics), and the combination of these properties makes them unique. So, we will further consider the evolving agent populations to create a definition for their self-organised \emph{diversity}.}

\subsection{Evolving Agent Populations}

\added{The self-organised \emph{diversity} of an evolving agent population comes from the agent-sequences it \emph{evolves}, in response to the \emph{selection pressure}, seeded with agents and agent-sequences from the agent-pool of the habitat in which it is instantiated. The set of agents and agent-sequences available when seeding an evolving agent population is regulated over time by other evolving agent populations, instantiated in response to other user requests, leading to the death and migration of agents and agent-sequences, as well as the formation of new agent-sequence combinations. The seeding of existing agent-sequences provides a direction to accelerate the evolutionary process, and can also affect the self-organised \emph{diversity}; for example, if only a proportion of any available global optima is favoured. So, the set of agents available when seeding an evolving agent population provides potential for the self-organised \emph{diversity}, while the \emph{selection pressure} of a \emph{user request} provides a constraining factor on this \emph{potential}. Therefore, the \emph{optimality} of the self-organised \emph{diversity} of an evolving agent population is \emph{relative} to the \emph{selection pressure} of the user request for which it was instantiated.}

While we could measure the self-organised \emph{diversity} of individual evolving agent populations, or even take a random sampling, it will be more informative to consider their \emph{collective} self-organised \emph{diversity}. \changed{Additionally, given that the Digital Ecosystem is required to support a range of user behaviour, we can consider the \emph{collective} self-organised \emph{diversity} of the evolving agent populations relative to the \emph{global} user request behaviour.} So, when varying a behavioural property of the user requests according to some distribution, we would expect the corresponding property of the evolving agent populations to follow the same distribution. \changed{While not intending to prescribe the expected user behaviour of the Digital Ecosystem, we do wish to investigate whether the Digital Ecosystem can adapt to a range of user behaviour.} So, we will consider Uniform, Gaussian (Normal) and Power Law distributions for the parameters of the user request behaviour. The Uniform distribution will provide a control, while the Normal (Gaussian) distribution will provide a reasonable assumption for the behaviour of a large group of users, and the Power Law distribution will provide a relatively extreme variation in user behaviour.

\changed{We therefore simulated the Digital Ecosystem, varying aspects of the user behaviour according to different distributions, and measuring the related aspects of the evolving agent populations.} This consisted of a mechanism to vary the user request properties of \emph{length} and \emph{modularity} (number of attributes per atomic service), according to Uniform, Gaussian (normal) and Power Law distributions, and a mechanism to measure the corresponding application (agent-sequence) properties of \emph{size} and \emph{number of attributes per agent}. \changed{For statistical significance each scenario (experiment) was averaged from ten thousand simulation runs.} We expect it will be obvious whether the \emph{observed} behaviour of the Digital Ecosystem matches the \emph{expected} behaviour from the user base. \changed{Nevertheless, we will also implement a chi-squared ($\chi^2$) test to confirm if the observed behaviour (distribution) of the agent-sequence properties matched the expected behaviour (distribution) from the user request properties.}

\section{Simulation and Results}

\added{We simulated the Digital Ecosystem, using our simulation from section 2 and \citep{thesis}. Including simulated populations of agent-sequences, $[A_1, A_1, A_2,$ $...]$, which were evolved to solve user requests, seeded with agents from the \emph{agent-pool} of 20 agents from the habitats in which they were instantiated.} \changed{A dynamic population size was used to ensure exploration of the available combinatorial search space, increasing with the average size of the population's agent-sequences.} The optimal combination of agents (agent-sequence) was evolved to the user request $R$, by an artificial \emph{selection pressure} created by a \emph{fitness function} generated from the user request $R$. An individual (agent) of the population consisted of a set of attributes, ${a_1, a_2, ...}$, and a user request consisted of a set of required attributes, ${r_1, r_2, ...}$. So, the \emph{fitness function} for evaluating an individual agent-sequence $A$, relative to a user request $R$, was
\begin{equation}
fitness(A,R) = \frac{1}{1 + \sum_{r \in R}{|r-a|}},
\label{ff}
\end{equation}
where $a$ is the member of $A$ such that the difference to the required attribute $r$ was minimised. \added{The abstract agent descriptions was based on existing and emerging technologies for \emph{semantically capable} \aclp{SOA} \citep{SOAsemantic}, such as the \acs{OWL-S} semantic markup for web services \citep{martin2004bsw}. We simulated an agent's \emph{semantic description} \setCap{with an abstract representation consisting of a set of}{as3} attributes, to simulate the properties of a \emph{semantic description}. Each \setCap{attribute representing a \emph{property} of the \emph{semantic description}, ranging between one and a hundred.}{agentSemantic2} \setCap{Each simulated agent was initialised with a semantic description}{as4} of between three and six attributes, which would then evolve in number and content.} 

Equation \ref{ff} was used to assign \emph{fitness} values between 0.0 and 1.0 to each individual of the current generation of the population, directly affecting their ability to replicate into the next generation. \changed{The evolutionary computing process was encoded with a low mutation rate, fixed selection pressure and non-trapping fitness function (i.e. did not get trapped at local optima\added{\footnote{\added{These constraints can be considered in abstract using the metaphor of the \emph{fitness landscape}, in which individuals are represented as solutions to the problem of survival and reproduction \citep{wright1932}. All possible solutions are distributed in a space whose dimensions are the possible properties of individuals. An additional dimension, height, indicates the relative fitness (in terms of survival and reproduction) of each solution. The fitness landscape is envisaged as a rugged, multidimensional landscape of hills, mountains, and valleys, because individuals with certain sets of properties are \emph{fitter} than others \citep{wright1932}. Here the ruggedness of the fitness landscape is not so severe, relative to the population diversity (population size and mutation rate), to prevent the evolving population from progressing to the global optima of the fitness landscape.}}}).} The type of selection used \emph{fitness-proportional} and \emph{non-elitist}, \emph{fitness-proportional means that the \emph{fitter} the individual the higher its probability of} surviving to the next generation. \emph{Non-elitist} means that the best individual from one generation was not guaranteed to survive to the next generation; it had a high probability of surviving into the next generation, but it was not guaranteed as it might have been mutated. \changed{\emph{Crossover/recombination} was then applied to a randomly chosen 10\% of the surviving population.} \emph{Mutations} were then applied to a randomly chosen 10\% of the surviving population; \emph{point mutations} were randomly located, consisting of \emph{insertions} (an agent was inserted into an agent-sequence), \emph{replacements} (an agent was replaced in an agent-sequence), and \emph{deletions} (an agent was deleted from an agent-sequence). \changed{The issue of bloat was controlled by augmenting the \emph{fitness function} with a \emph{parsimony pressure} to bias the search to smaller agent-sequences, evaluating larger than average agent-sequences with a reduced \emph{fitness}, and therefore providing a dynamic control limit adaptive to the average size of the individuals of the ever-changing evolving agent populations.}

\added{Our simulation included our extended Physical Complexity and Efficiency, which required implementing the $C_V$ of (\ref{newComplexity}), the $\ell_V$ of (\ref{clustersSampleSize}) and the $H_V$ of (\ref{perSiteVLP}) for the per-site entropies. The Efficiency $E_{c}$ (\ref{efficiencyMultiple}), for populations with clusters, was also implemented in the simulation. It also included our extended Chli-DeWilde stability and \emph{degree of instability}, which required calculating $p_{\bx}^{(t)}$ of (\ref{eq5.1}) to estimate the stability, and $p_\bx^\infty$ of (\ref{eq2}) to prove the existence of $p_\bx^\infty \neq p_\by^\infty$ from (\ref{eq3}). The \emph{degree of instability}, $d_{ins}$ of (\ref{eq6}), was also implemented in the simulation.}

\added{For the diversity experiments we included a way to vary aspects of the user behaviour according to different distributions, and a way to measure the related aspects of the evolving agent populations. This consisted of a mechanism to vary the user request properties of \emph{length} and \emph{modularity}, according to Uniform, Gaussian (normal) and Power distributions, and a mechanism to measure the corresponding agent(-sequence) properties of \emph{length} and \emph{number of attributes}. For statistical significance each scenario (experiment) will be averaged from ten thousand simulation runs. We expect it will be obvious whether the \emph{observed} behaviour of the Digital Ecosystem matches the \emph{expected} behaviour from the user base. Nevertheless, we will also implement a chi-squared ($\chi^2$) test to determine if the observed behaviour (distribution) of the agent(-sequence) properties matches the expected behaviour (distribution) from the user request properties.}

\subsection{Complexity}

Figure \ref{phycom} shows, for a typical evolving agent population, the Physical Complexity $C_V$ (\ref{newComplexity}) for \aclp{vls} and the \emph{maximum fitness} $F_{max}$ over the generations. It shows that the fitness and our extended Physical Complexity; both increase over the generations, synchronised with one another, until generation 160 when the \emph{maximum fitness} tapers off more slowly than the Physical Complexity. \changed{At this point the optimal length for the sequences is reached within the simulation, and so the advent of new fitter sequences (of the same of similar length) creates only minor fluctuations in the Physical Complexity, while having a more significant effect on the \emph{maximum fitness}.} \added{It increases over the generations because of the increasing information being stored, with the sharp increases occurring when the effective length $\ell_{V}$ of the population increases. The temporary decreases, such as the one beginning at generation 138, are preceded by the advent of a new \emph{fitter} mutant, as indicated by a corresponding sharp increase in the \emph{maximum fitness} in the immediately preceding generations, which temporarily disrupt the self-organised complexity of the population, until this new fitter mutant becomes dominant and leads to a new higher level of self-organised complexity.} \changed{The similarity of the graph in Figure \ref{phycom} to the graphs in \citep{adami20002} confirms that the Physical Complexity measure has been successfully extended to \aclp{vls}.}

\tfigure{width=\textwidth}{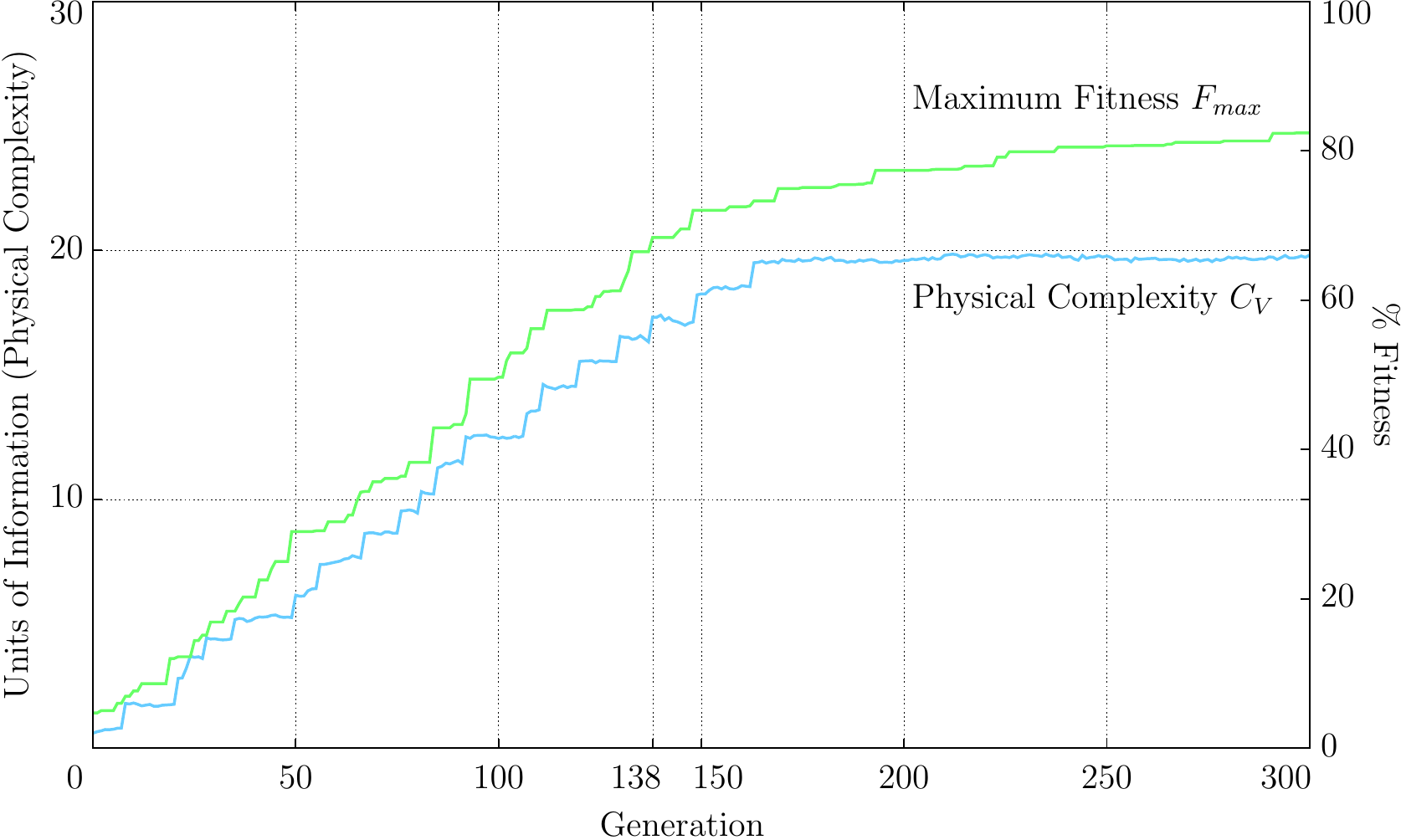}{graph}{\changed{Graph of Physical Complexity and Maximum Fitness}}{The Physical Complexity for \aclp{vls} increases over the generations, showing short-term decreases as expected, such as at generation 138.}{-2mm}{!t}{}{}

\subsubsection{Efficiency}

\changed{Figure \ref{newphycomvis} is a visualisation of the simulation, showing two alternate populations run for a thousand generations, with the one on the left from Figure \ref{phycom} run under normal conditions, while the one on the right was run with a non-discriminating selection pressure.} Each multi-coloured line represents an agent-sequence, while each colour represents an agent (site). \changed{\setCap{The visualisation shows that our Efficiency $E$ accurately measures the self-organised complexity of the two populations.}{largeVisCap}} \added{It also shows significant variation in the population run under normal conditions, as the evolutionary computing process creates the opportunity to find fitter (better) sequences, providing potential to avoid getting trapped at local optima.}

\tfigure{width=\textwidth}{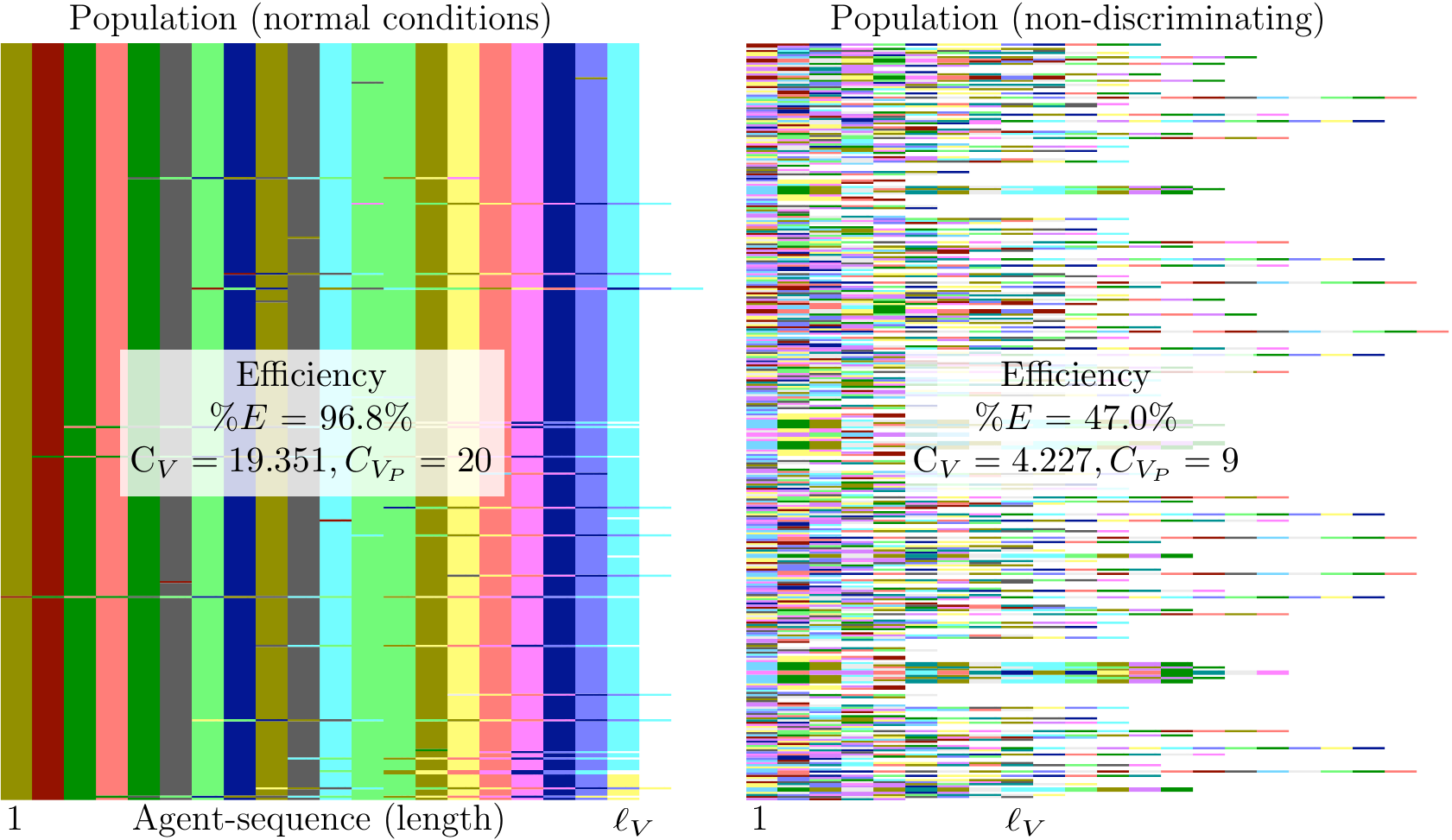}{graffle}{Visualisation of Evolving Agent Populations at the 1000th Generation}{\changed{The population on the left, from Figure \ref{phycom}, was run under normal conditions, while the other was run with a non-discriminating selection pressure.}}{-2mm}{!h}{}{}

Figure \ref{efficiency} shows the Efficiency $E$ (\ref{efficiencyEQ}), over the generations, for the population from Figure \ref{phycom}. \changed{\setCap{The Efficiency tends to a maximum of one, indicating that the population consists of one cluster}{graph32cap}, which is confirmed by the visualisation of the population in Figure \ref{newphycomvis} (left).} \added{The significant decreases that occurred in the Efficiency, reducing in magnitude and frequency over the generations, came from mirroring the fluctuations that occurred in the complexity $C_V$, because the Efficiency $E$ (\ref{efficiencyEQ}) is the complexity $C_V$ (\ref{newComplexity}) over the complexity potential $C_{V_{P}}$(\ref{potential}). These falls are caused by the creation of fitter (better) mutants within the population, which eventually become the dominant genotype, but during the process causes the Physical Complexity and the Efficiency to fall in the short-term.}

\tfigure{width=\textwidth}{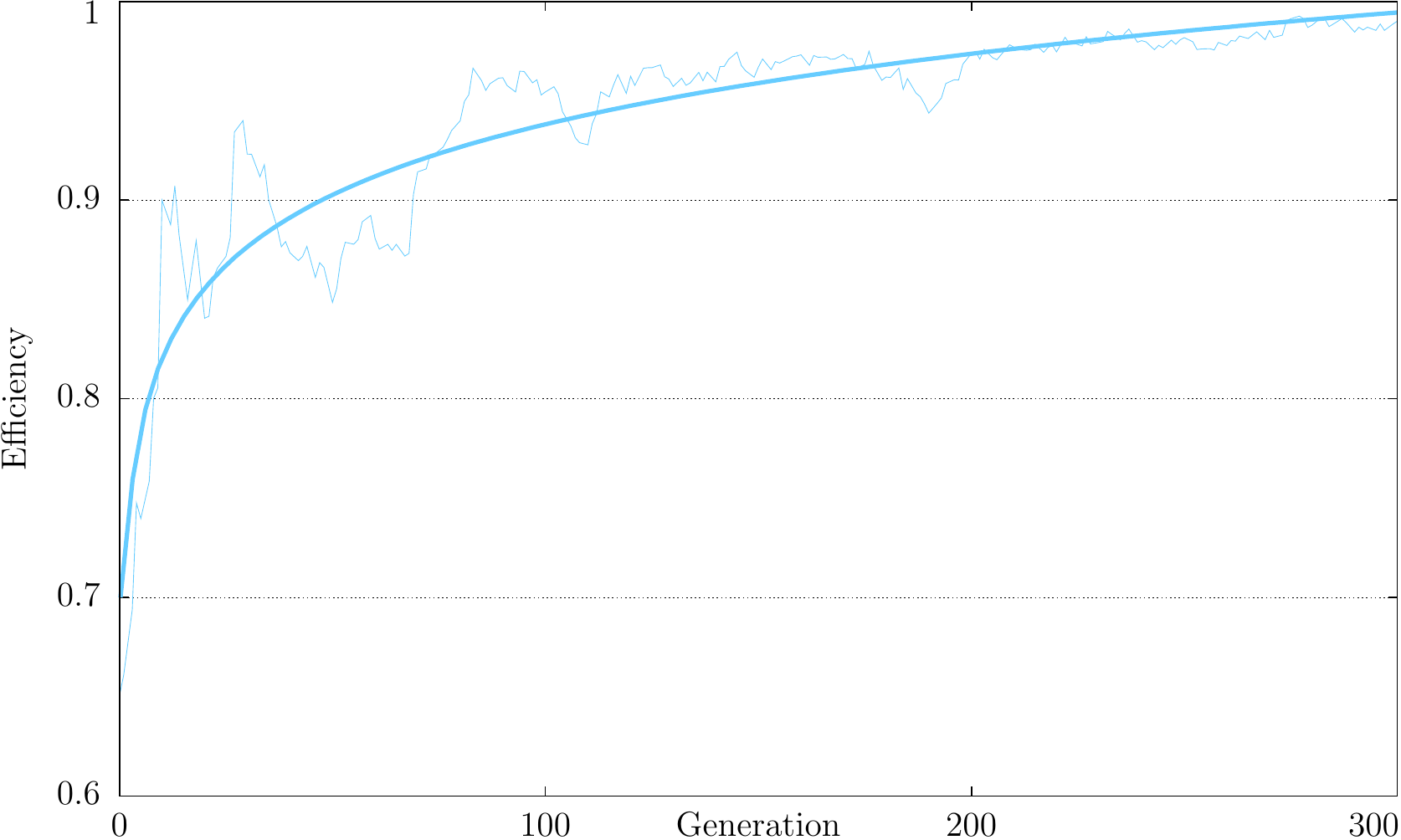}{graph}{Graph of Population Efficiency over the Generations for the population from Figure \ref{phycom}}{\getCap{graph32cap}.}{-2mm}{}{}{}

\subsubsection{\added{Clustering}}

\tfigure{width=\textwidth}{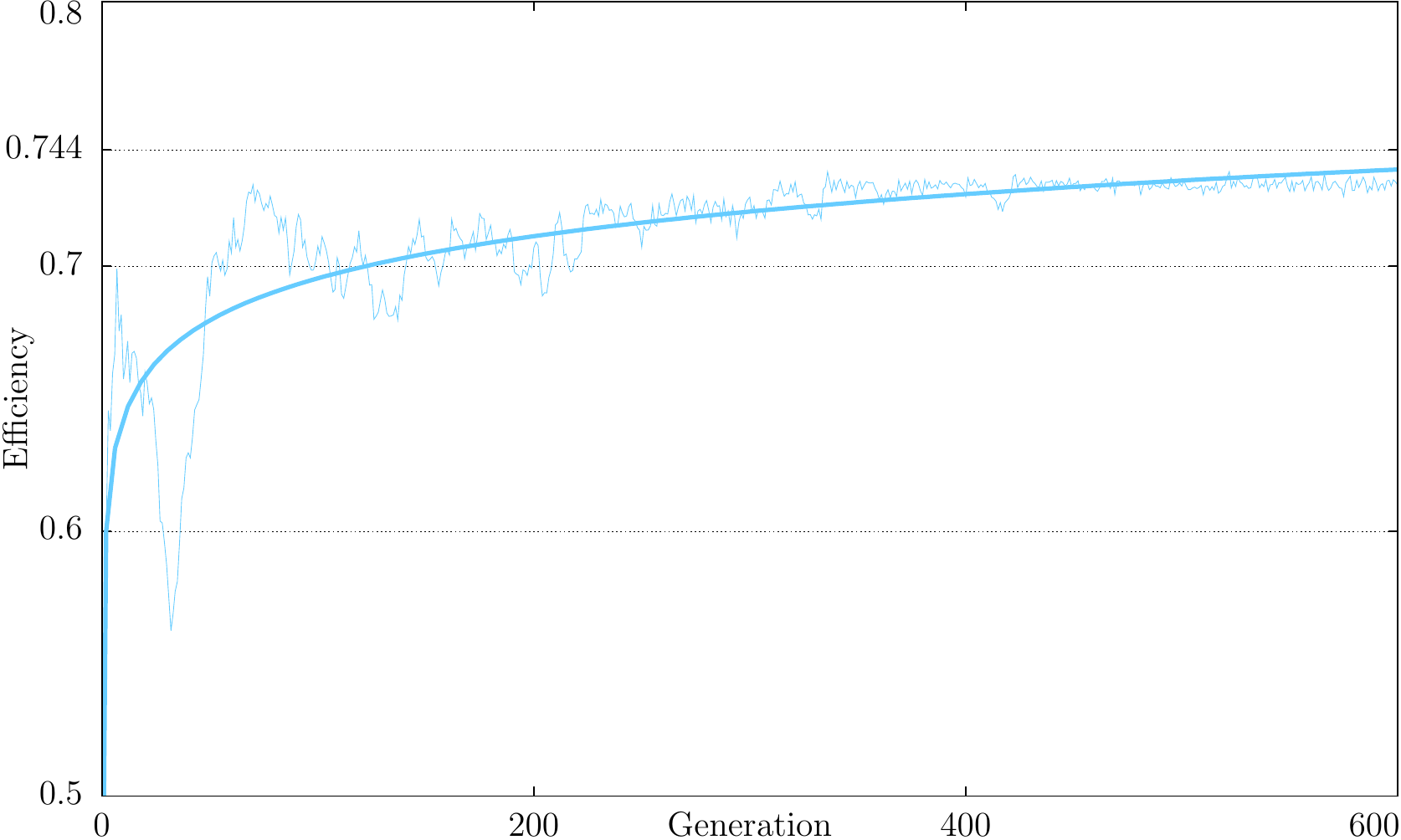}{graph}{\added{Graph of the Clustering Coefficient}}{\added{The Efficiency oscillated to 0.744, as expected from (\ref{calcNumClusters2}) given the alphabet size was fifteen, $|D|$=15, and the number of clusters was two, $|T|$=2, indicating more than one cluster.}}{-2mm}{!t}{}{}

\tfigure{width=\textwidth}{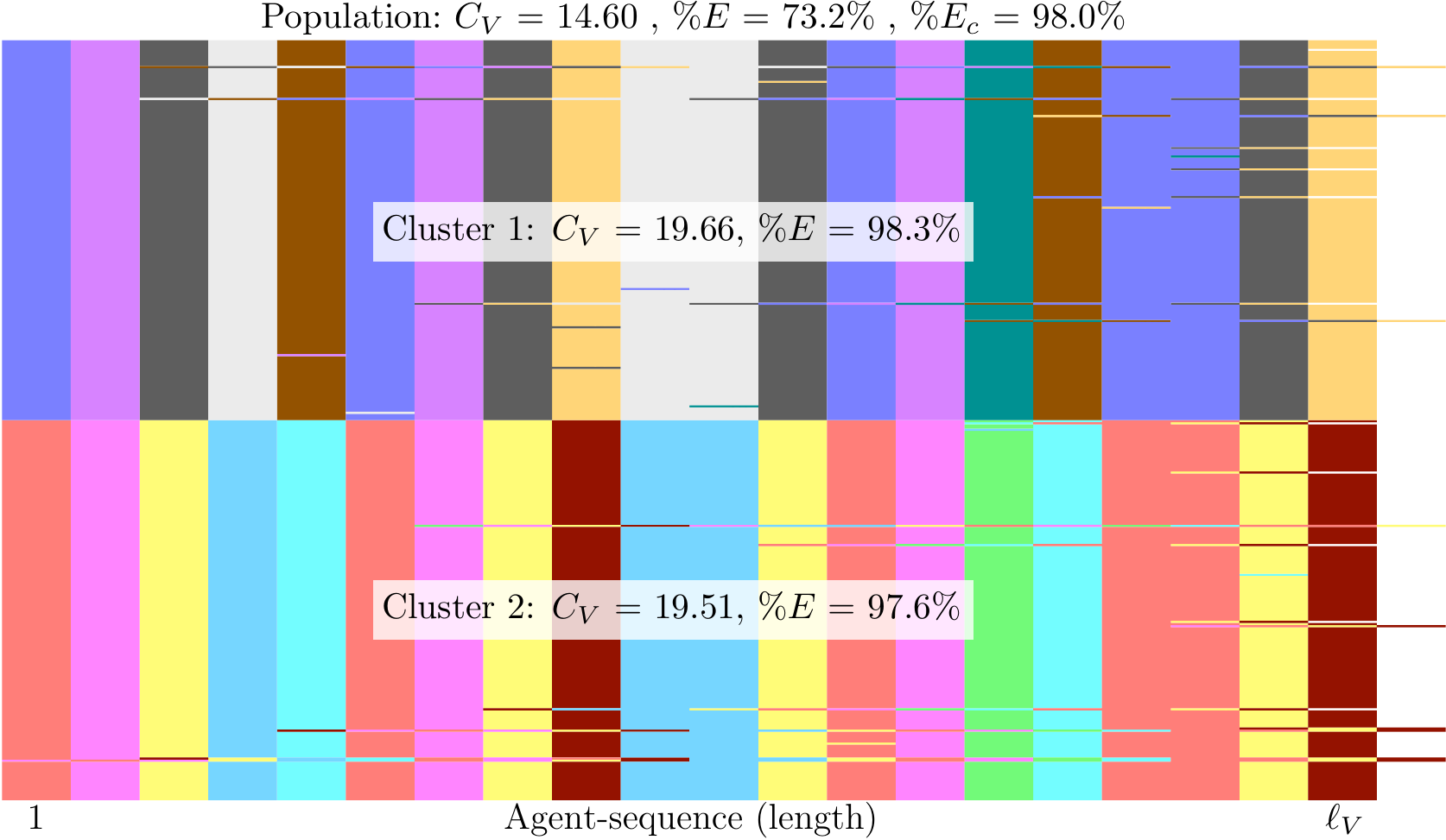}{graffle}{\added{Visualisation of Clusters in an Evolving Agent Population at the 1000th Generation}}{\added{The \getCap{visClustersCap}.}}{-2mm}{!b}{}{}

\added{To further investigate the self-organisation of evolving agent populations, we simulated a typical population with a multi-objective \emph{selection pressure} that had two independent global optima, and so the potential to support two \emph{pure clusters} (each cluster using a unique subset of the alphabet $D$). The graph in Figure \ref{coefficient} shows the Efficiency $E$ over the generations acting as a \emph{clustering coefficient}, oscillating \setCap{around the included best fit curve, quite significantly at the start, and then decreasing as the generations progressed.}{graph4cap} The Efficiency tended to 0.744, as expected from (\ref{calcNumClusters2}) given the alphabet size was fifteen, $|D|$=15, and the number of clusters was two, $|T|$=2. The tending itself indicated clustering, while the value it tended to indicated, as expected, the presence of two clusters in the population. The initial severe oscillations were caused by the creation and spread of fitter \emph{longer} mutants (agent-sequences) in the population, causing the Physical Complexity and therefore the Efficiency to fluctuate significantly. A visualisation of the population is shown in Figure \ref{newCluster}, in which the \setCap{agent-sequences were grouped to show the two clusters}{visClustersCap}. As \setCap{expected from (\ref{defineCluster}) each cluster had a much higher Physical Complexity and Efficiency compared to the population as a whole. However, the Efficiency $E_{c}$}{visClusters2Cap} is immune to the clusters and therefore \setCap{calculated the}{visClusters3Cap} self-organisation of the population correctly.}

\subsubsection{\added{Summary}}

\changed{Collectively, the experimental results confirm that Physical Complexity has been successfully extended to evolving agent populations. Most significantly, Physical Complexity has been reformulated algebraically for populations of \aclp{vls}, which we have confirmed experimentally through simulations. Our \emph{Efficiency} definition provides a macroscopic value to characterise the level of \emph{complexity}. Furthermore, the \emph{clustering coefficient} defined by the tending of the Efficiency, not only indicates clustering, but can also distinguish between a single cluster population and a population with clusters. The number of clusters can even be determined, for \emph{pure clusters}, from the value to which the \emph{clustering coefficient} tends. Combined, this allows the \emph{Efficiency} $E_{c}$ definition to provide a normalised \emph{universally applicable} macroscopic value to characterise the \emph{complexity} of a population, independent of clustering, atomicity, length (variable or same), and size.}

\subsection{Stability}

Our evolving agent population (a \ac{MAS} with evolutionary dynamics) is stable if the distribution of the limit probabilities exists and is non-uniform, as defined by equations (\ref{eq2}) and (\ref{eq3}). \changed{The simplest scenario is a typical evolving agent population with a single global optimal solution, which is stable if there are at least two macro-states with different limit occupation probabilities.} We shall consider the \emph{maximum macro-state} $M_{max}$ and the \emph{sub-optimal macro-state} $M_{half}$. Where the states of the macro-state $M_{max}$ each possess at least one individual with global maximum fitness,
\begin{equation*}
p_{\bone}^\infty = lim_{t \rightarrow \infty} p_{\bone}^{(t)} = 1,
\end{equation*}
while the states of the macro-state $M_{half}$ each possess at least one individual with a fitness equal to \emph{half} of the global maximum fitness,
\begin{equation*}
p_{\bfif}^\infty = lim_{t \rightarrow \infty} p_{\bfif}^{(t)} = 0,
\end{equation*}
thereby fulfilling the requirements of equations (\ref{eq2}) and (\ref{eq3}). \changed{The \emph{sub-optimal macro-state} $M_{half}$, having a lower fitness, we predict to be seen earlier in the evolutionary process before disappearing as higher fitness macro-states are reached.} The system $S$ will take longer to reach the \emph{maximum macro-state} $M_{max}$, but once it does will likely remain, leaving only briefly depending on the strength of the mutation rate, as the \emph{selection pressure} is \emph{non-elitist}.

\changed{A value of $t=1000$ was chosen to represent $t=\infty$ experimentally, because the simulation has often been observed to reach the \emph{maximum macro-state} $M_{max}$ by 500 generations.} Therefore, the probability of the system $S$ being in the \emph{maximum macro-state} $M_{max}$ at the thousandth generation is expected to be one, $p^{1000}_{\bone} = 1$. Furthermore, the probability of the system being in the \emph{sub-optimal macro-state} $M_{half}$ at the thousandth generation is expected to be zero, $p^{1000}_{\bfif} = 0$. \added{We can therefore conclude that our extended Chli-DeWilde stability accurately models the stability over time of evolving agent populations.}

\tfigure{width=\textwidth}{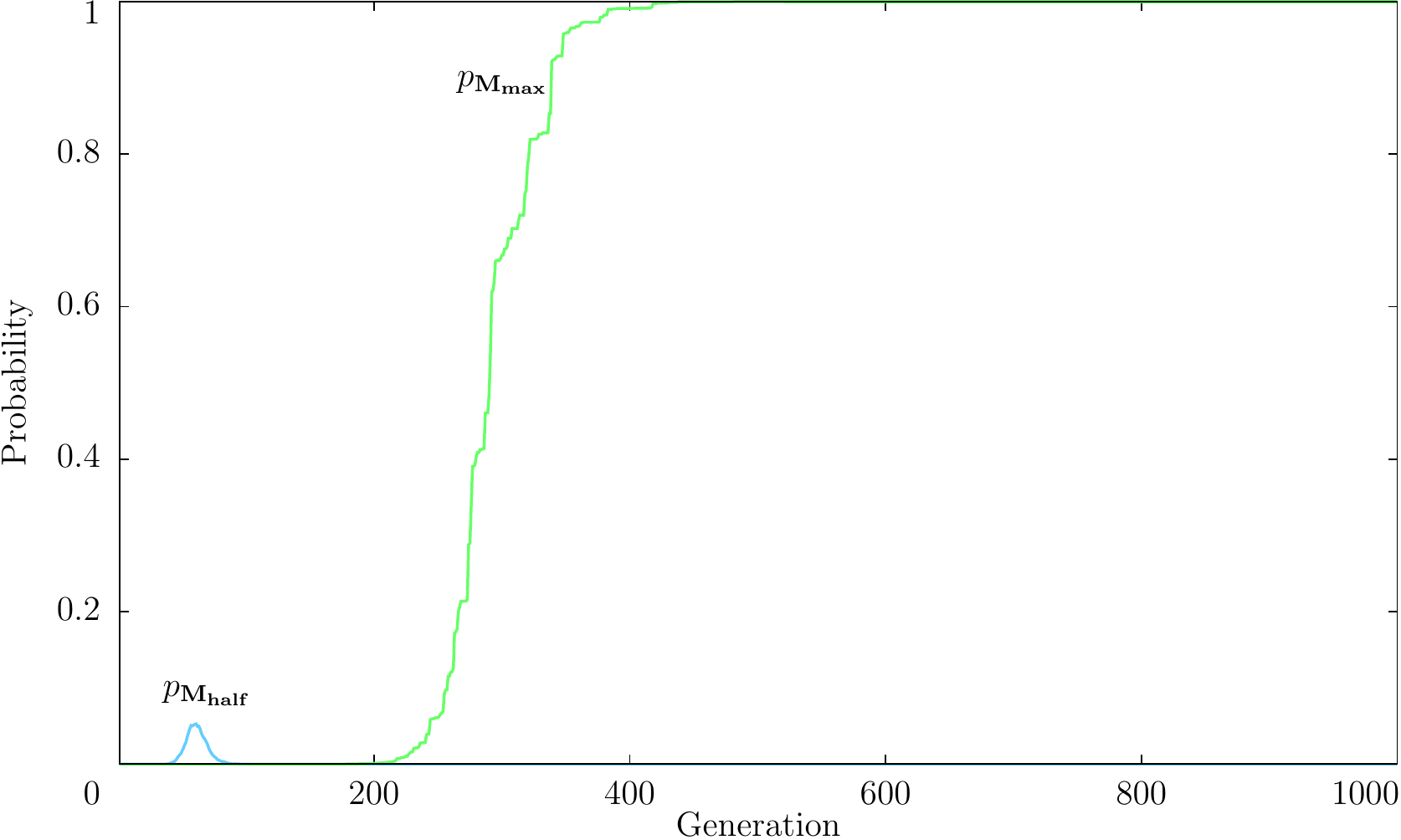}{graph}{Graph of the Probabilities of the Macro-States}{$M_{max}$ and $M_{half}$ at each Generation: The system $S$, a typical evolving agent population, was \getCap{graphCap}}{-2mm}{!b}{}{}

\changed{Figure \ref{macrostates} shows, for a typical evolving agent population, a graph of the probability as defined by equation (\ref{eq5.1}) of the \emph{maximum macro-state} $M_{max}$ and the \emph{sub-optimal macro-state} $M_{half}$ at each generation, averaged from ten thousand simulation runs to provide statistical significance.} The behaviour of the simulated system $S$ was as expected, being \setCap{in the \emph{maximum macro-state} $M_{max}$ only after generation 178 and always after generation 482.}{graphCap} It was also observed being in the \emph{sub-optimal macro-state} $M_{half}$ only between generations 37 and 113, with a maximum probability of 0.053 (3 d.p.) at generation 61, and was such because the evolutionary path (state transitions) could avoid visiting the macro-state. \changed{As we expected the probability of being in the \emph{maximum macro-state} $M_{max}$ by the thousandth generation was one, $p^{1000}_{\bone} = 1$, and so the probability of being in any other macro-state, including the \emph{sub-optimal macro-state} $M_{half}$, by the thousandth generation was zero, $p^{1000}_{\bfif} = 0$.}

A visualisation for the state of a typical evolving agent population at the thousandth generation is shown in Figure \ref{vis62}, with each line representing an agent-sequence and each colour representing an agent, with the identical agent-sequences grouped for clarity. \changed{It shows that the evolving agent population reached the \emph{maximum macro-state} $M_{max}$ and remained there, but as we expected never reached the \emph{maximal state} of the \emph{maximum macro-state}, where all the agent-sequences are identical and have maximum fitness, indicated by the lack of total uniformity in Figure \ref{vis62}.} This was expected, because of the mutation (noise) within the evolutionary process, which is necessary to create the opportunity to find fitter (better) sequences and potentially avoid getting trapped at any local optima that may be present. \added{We can therefore conclude that the macro-state interpretation of our extended Chli-DeWilde stability accurately models the state-space of evolving agent populations.}

\tfigure{width=\textwidth}{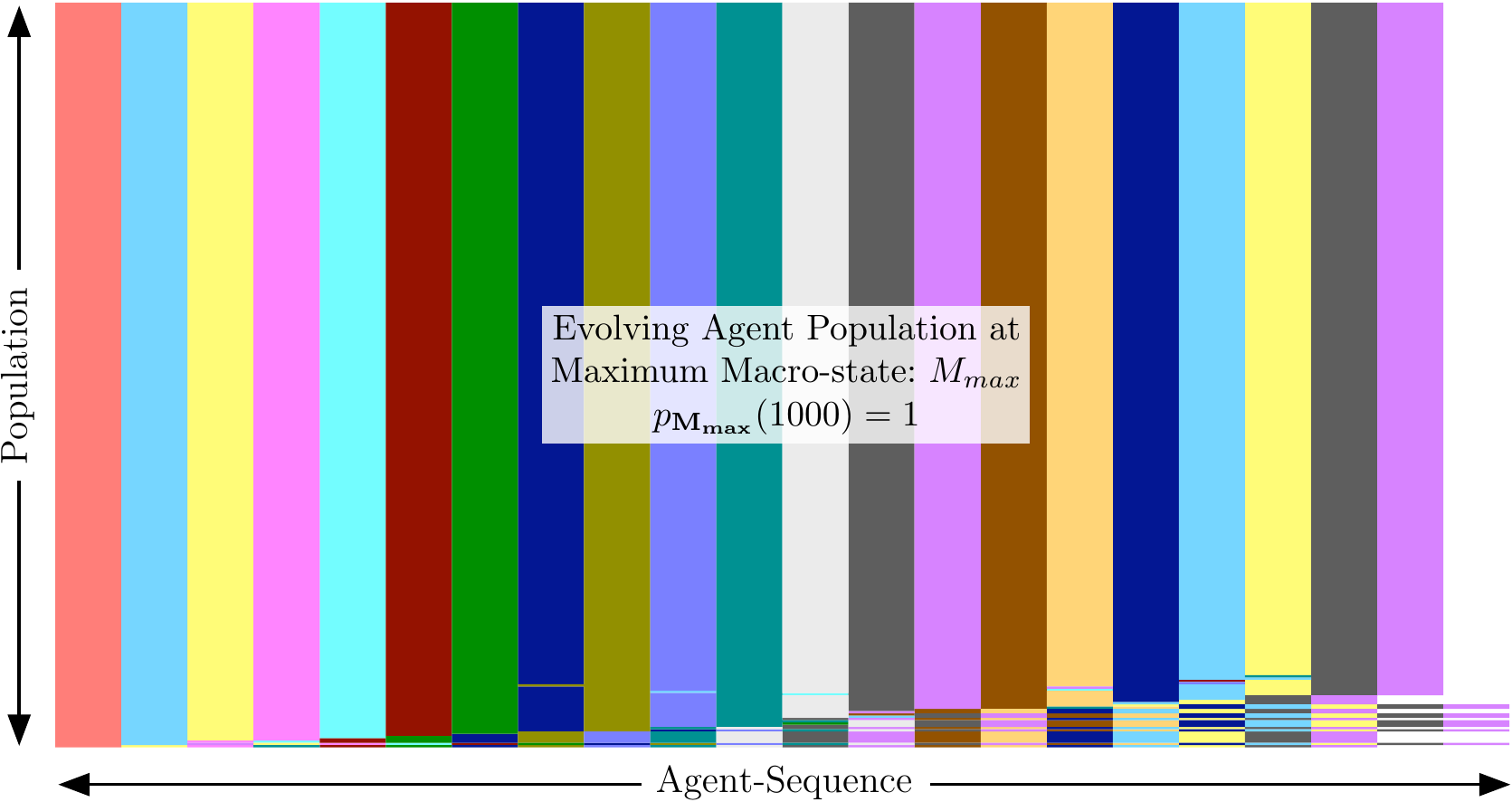}{graffle}{Visualisation of an Evolving Agent Population at the 1000th Generation}{\changed{The population consists of multiple agent-sequences, with each line representing an agent-sequence, and so each colour representing an individual agent.}}{-3mm}{!h}{}{}

\subsection{Degree of Instability}

\changed{Given that our simulated evolving agent population is stable, as defined by equations (\ref{eq2}) and (\ref{eq3}), we can determine its \emph{degree of instability} as defined by equation (\ref{eq6}).} So, calculated from its limit probabilities, the \emph{degree of instability} was

\vspace{-5mm}

\begin{eqnarray}
d_{ins} = H(p^{1000}) &=& -\sum\limits_{\bx}p^{1000}_\bx log_{N}(p^{1000}_\bx) \nonumber \\ 
 &=& -1log_{N}(1) \nonumber \\
 &=& 0, \nonumber
\label{result2}
\end{eqnarray}

where $t=1000$ is an effective estimate for $t=\infty$, as explained earlier. \changed{The result was as we expected because the \emph{maximum macro-state} $M_{max}$ by the thousandth generation was one, $p^{1000}_{\bone} = 1$, and so the probability of being in the other macro-states by the thousandth generation was zero.} The system therefore shows no instability, as there is no entropy in the occupied macro-states at infinite time. \added{We can therefore conclude that the \emph{degree of instability} of our extended Chli-DeWilde stability can provide a macroscopic value to characterise the \emph{level of stability} of evolving agent populations.}

\subsection{Stability Analysis}

\tfigure{width=\textwidth}{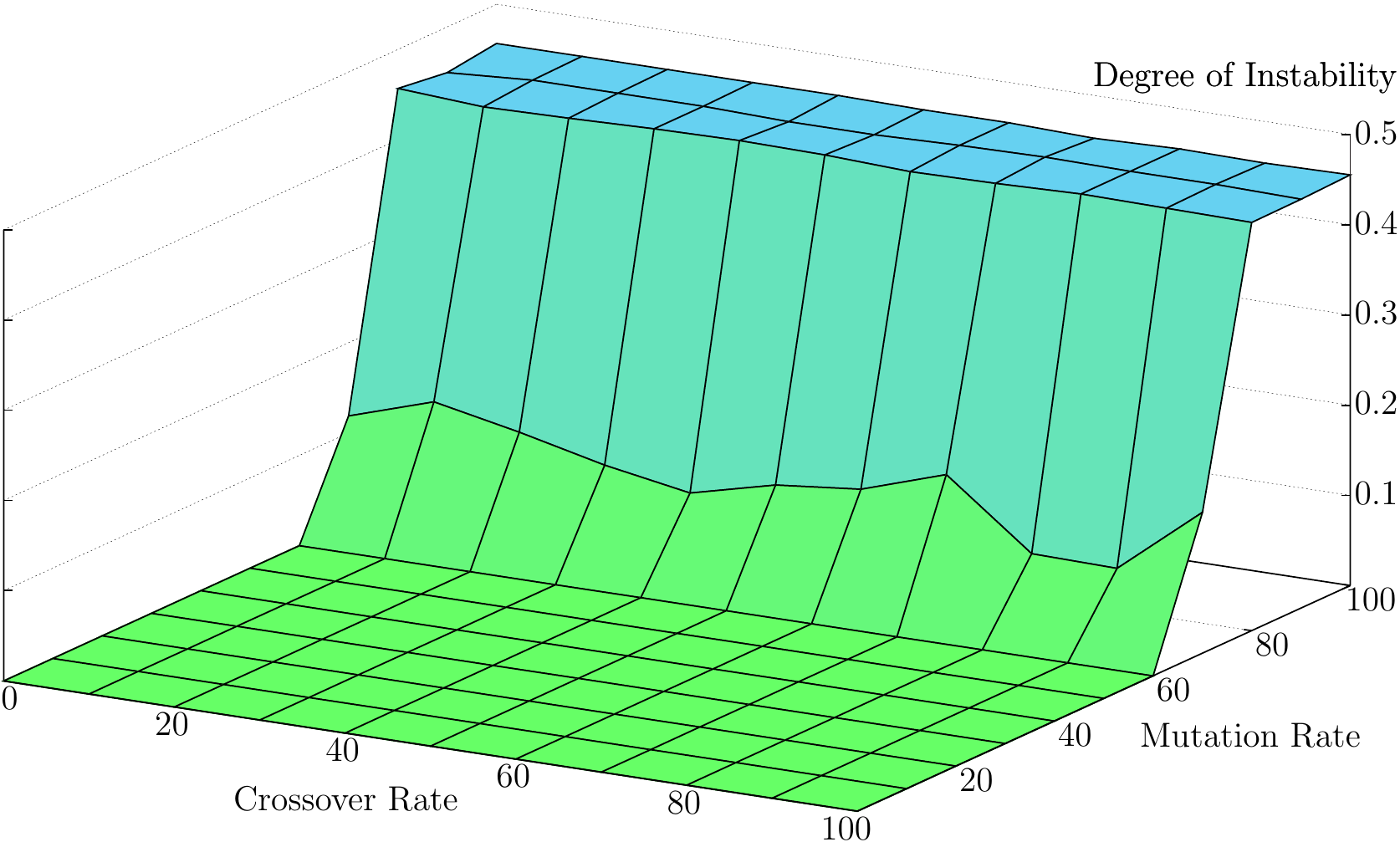}{graph}{Graph of Stability with Different Mutation and Crossover Rates}{\getCap{aScap}.}{-3mm}{!b}{}{}

\changed{We then performed a \emph{stability analysis} (akin to a \emph{sensitivity analysis} \citep{cacuci2003sau}) of a typical evolving agent population, varying its key parameters while measuring its stability.} We varied the mutation and crossover rates from 0\% to 100\% in 10\% increments \added{to provide a sufficient density of measurements to identify any trends that might be present}, calculating the \emph{degree of instability}, $\delta$ from (\ref{eq6}), at the thousandth generation. These \emph{degree of instability} values were averaged over 10 000 simulation runs \added{to ensure statistical significance}, and graphed against the mutation and crossover rates in Figure \ref{stabilityAnalysis}. \changed{It showed that the crossover rate had little effect on the stability of the simulated evolving agent population, whereas the mutation rate did have a significant affect on the stability.} \setCap{With the mutation rate under or equal to 60\%, the evolving agent population showed no instability}{aScap}, with $d_{ins}$ values equal to zero as the system $S$ was always in the same macro-state $M$ at infinite time, independent of the crossover rate. \changed{With the mutation rate above 60\% the was a significant increase in instability, with the system being in one of several different macro-states at infinite time; with a mutation rate of 70\% its was still very stable, having low $d_{ins}$ values ranging between 0.08 and 0.16, but once the mutation rate was 80\% or greater it became quite unstable, shown by high $d_{ins}$ values nearing 0.5.}

\changed{As one would expect, an \emph{extremely} high mutation rate has a destabilising effect on the \emph{stability} of an evolving agent population.} \added{Also, as expected} \changed{t}he crossover rate had only a minimal effect, because variation from \emph{crossover} was limited when the population had \emph{matured}, consisting of agent-sequences identical or very similar to one another. \changed{It should also be noted that the stability is different to performance, because although showing no instability with mutation rates below 60\% (inclusive), it only reached the \emph{maximum macro-state} $M_{max}$ with a mutation rate of 10\% or above, while at 0\% it was stable at a sub-optimal macro-state.} \added{We can therefore conclude that the \emph{degree of instability} of our extended Chli-DeWilde stability can used of perform stability analyses (similar to a \emph{sensitivity analyses} \citep{cacuci2003sau}) of evolving agent populations.}

\subsubsection{Summary}

\added{Collectively, the experimental results confirm that Chli-Dewilde stability has been successfully extended to evolving agent populations, while our definition for the \emph{degree of instability} provides a macroscopic value to characterise the level of \emph{stability}.}

\subsection{Diversity}

\subsubsection{User Request Length}

\changed{We started by varying the \emph{user request length} according to the available distributions, expecting the size of the corresponding applications (agent-sequences) to be distributed according to the length of the user requests, i.e. the longer the user request, the larger the agent-sequence needed in response.}

We first applied the Uniform distribution as a control, and graphed the results in Figure \ref{urluniform}. \setCap{The \emph{observed} frequencies of the application (agent-sequence) size mostly matched the \emph{expected} frequencies}{urlCapUnifrom}, which was \emph{confirmed} with a $\chi^2$ test; with a \emph{null hypothesis} of \emph{no significant difference} and \emph{sixteen degrees of freedom}, the $\chi^2$ value was 2.588, below the critical 0.95 $\chi^2$ value of 7.962. 

\tfigure{width=\textwidth}{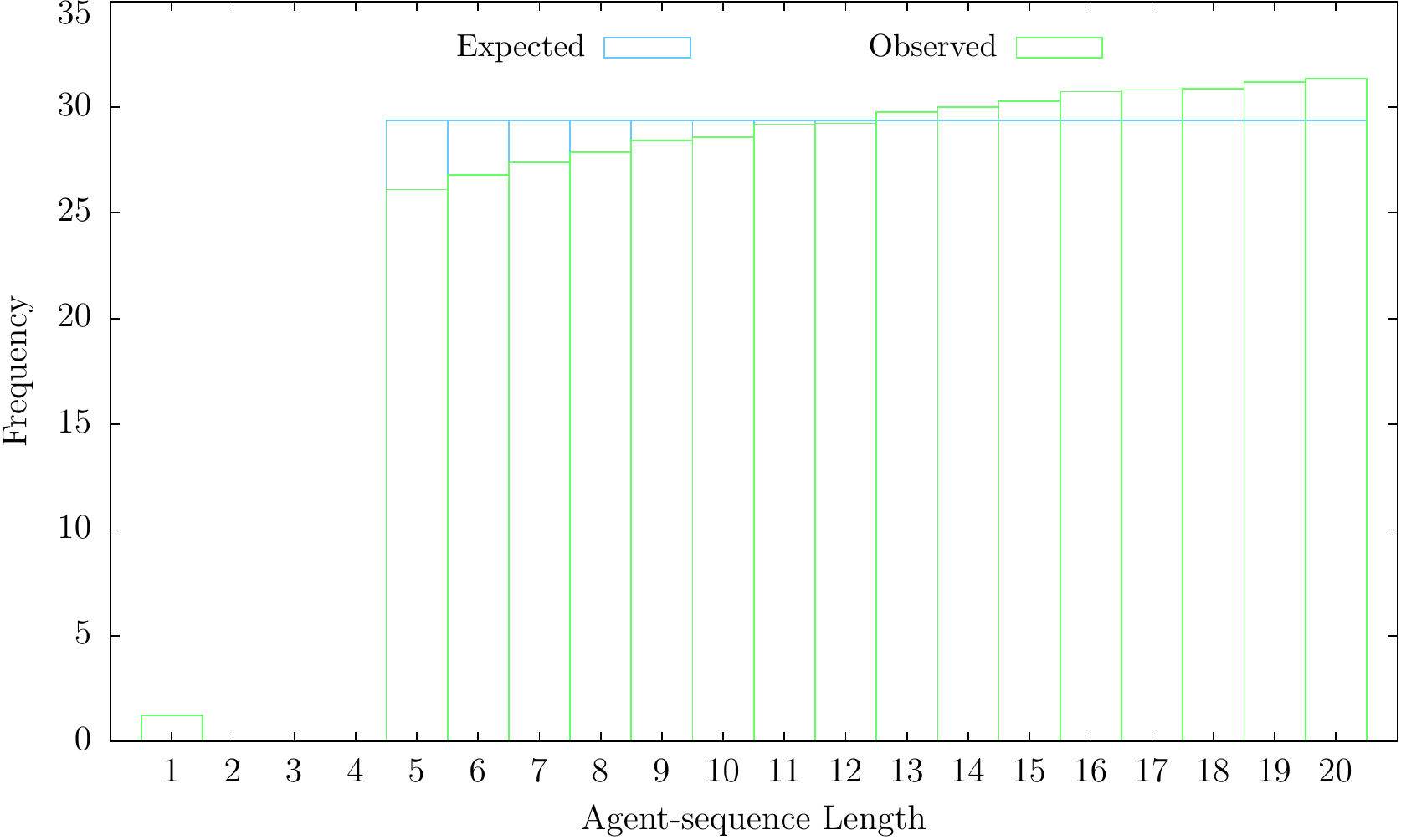}{graph}{Graph of Uniformly Distributed Agent-Sequence Length Frequencies}{\changed{\getCap{urlCapUnifrom},} \added{,which was \emph{confirmed} by a $\chi^2$ test; with a \emph{null hypothesis} of \emph{no significant difference} and \emph{sixteen degrees of freedom}, the $\chi^2$ value was 2.588, below the critical 0.95 $\chi^2$ value of 7.962.}}{-3mm}{!t}{0mm}{}

\tfigure{width=\textwidth}{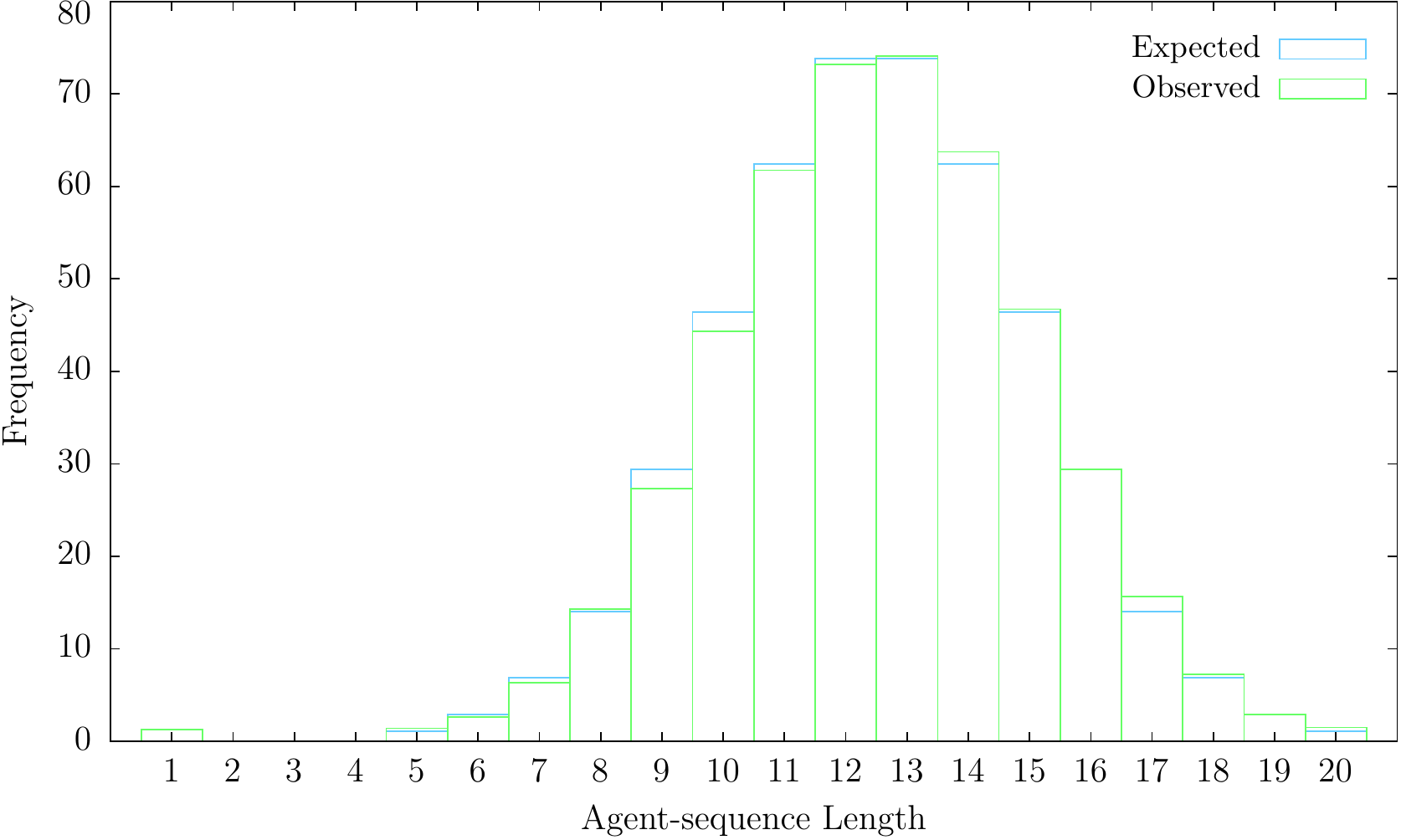}{graph}{Graph of Gaussian Distributed Agent-Sequence Length Frequencies}{\changed{\getCap{urlCapGaussian},} \added{which was confirmed by a $\chi^2$ test; with a \emph{null hypothesis} of \emph{no significant difference} and \emph{sixteen degrees of freedom}, the $\chi^2$ value was 2.102, below the critical 0.95 $\chi^2$ value of 7.962.}}{-3mm}{!b}{-10mm}{0mm}

We then applied the Gaussian distribution as a reasonable assumption for the behaviour of a large group of users, and graphed the results in Figure \ref{urlgaussian}. \setCap{The \emph{observed} frequencies of the application (agent-sequence) size matched the \emph{expected} frequencies with only minor variations}{urlCapGaussian}, which was confirmed by a $\chi^2$ test; with a \emph{null hypothesis} of \emph{no significant difference} and \emph{sixteen degrees of freedom}, the $\chi^2$ value was 2.102, below the critical 0.95 $\chi^2$ value of 7.962.

Finally, we applied the Power Law distribution to represent a relatively extreme variation in user behaviour, and graphed the results in Figure \ref{urlpower}. \setCap{The \emph{observed} frequencies of the application (agent-sequence) size matched the \emph{expected} frequencies with some variation}{urlpower}, which was confirmed by a $\chi^2$ test; with a \emph{null hypothesis} of \emph{no significant difference} and \emph{sixteen degrees of freedom}, the $\chi^2$ value was 5.048, below the critical 0.95 $\chi^2$ value of 7.962.

\tfigure{width=\textwidth}{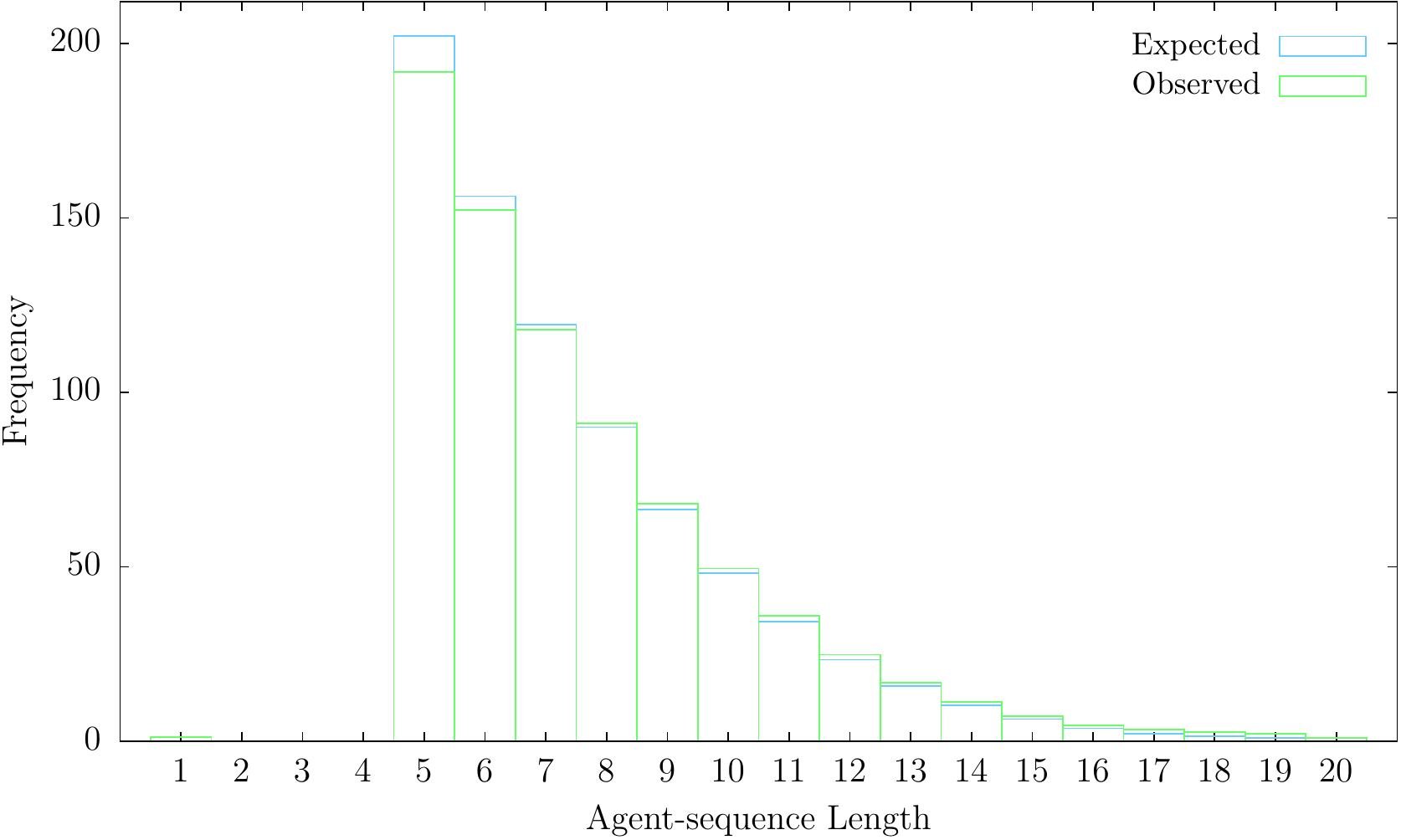}{graph}{Graph of Power Law Distributed Agent-Sequence Length Frequencies}{\changed{\getCap{urlpower},} \added{which was confirmed by a $\chi^2$ test; with a \emph{null hypothesis} of \emph{no significant difference} and \emph{sixteen degrees of freedom}, the $\chi^2$ value was 5.048, below the critical 0.95 $\chi^2$ value of 7.962.}}{-3mm}{!t}{0mm}{0mm}

There were a couple of minor discrepancies, similar to all the experiments. \changed{First, there were a small number of \emph{individual} agents at the thousandth time step, caused by the typical user behaviour of continuously creating new agents (services).} Second, while the chi-squared tests confirmed that there was no significant difference between the \emph{observed} and \emph{expected} frequencies of the application (agent-sequence) size, there was still a \emph{bias} to larger applications (solutions). \changed{Evident visually in the graphs of the experiments, and evident numerically in the chi-squared test of the Power Law distribution experiment as it favoured smaller agent-sequences}. The cause of this \emph{bias} was most likely some aspect of \emph{bloat}\added{\footnotemark} not fully controlled.

\footnotetext[3]{\added{When variable length representations of solutions are used, a well-known phenomenon arises, called \emph{bloat}, in which the individuals of an evolving population tend to grow in size without gaining any additional advantage \citep{langdon1997fcb}.}}

\subsubsection{User Request Modularity}

\changed{Next, we varied the \emph{user request modularity} (number of attributes per atomic service) according to the available distributions, expecting the \emph{sophistication} of the agents to be distributed according to the modularity of the user requests, i.e. the more complicated (in terms of modular non-reducible tasks) the user request, the more sophisticated (in terms of the number of attributes) the agents needed in response.}

We first applied the Uniform distribution as a control, and graphed the results in Figure \ref{urvuniform}. \setCap{The \emph{observed} frequencies for the number of agent attributes mostly matched the \emph{expected} frequencies}{urvunifromCap}, which was confirmed by a $\chi^2$ test; with a \emph{null hypothesis} of \emph{no significant difference} and \emph{ten degrees of freedom}, the $\chi^2$ value was 1.049, below the critical 0.95 $\chi^2$ value of 3.940.

\label{divmodexp}

We then applied the Gaussian distribution as a reasonable assumption for the behaviour of a large group of users, and graphed the results in Figure \ref{urvgaussian}. \setCap{The \emph{observed} frequencies for the number of agent attributes again followed the \emph{expected} frequencies, but there was variation}{urvgaussianCap} which led to a failed $\chi^2$ test; with a \emph{null hypothesis} of \emph{no significant difference} and \emph{ten degrees of freedom}, the $\chi^2$ value was 50.623, not below the critical 0.95 $\chi^2$ value of 3.940.

Finally, we applied the Power Law distribution to represent a relatively extreme variation in user behaviour, and graphed the results in Figure \ref{urvpower}. \setCap{The \emph{observed} frequencies for the number of agent attributes also followed the \emph{expected} frequencies, but there was variation}{urvpowerCap} which led to a failed $\chi^2$ test; with a \emph{null hypothesis} of \emph{no significant difference} and \emph{ten degrees of freedom}, the $\chi^2$ value was 61.876, not below the critical 0.95 $\chi^2$ value of 3.940.

\tfigure{width=\textwidth}{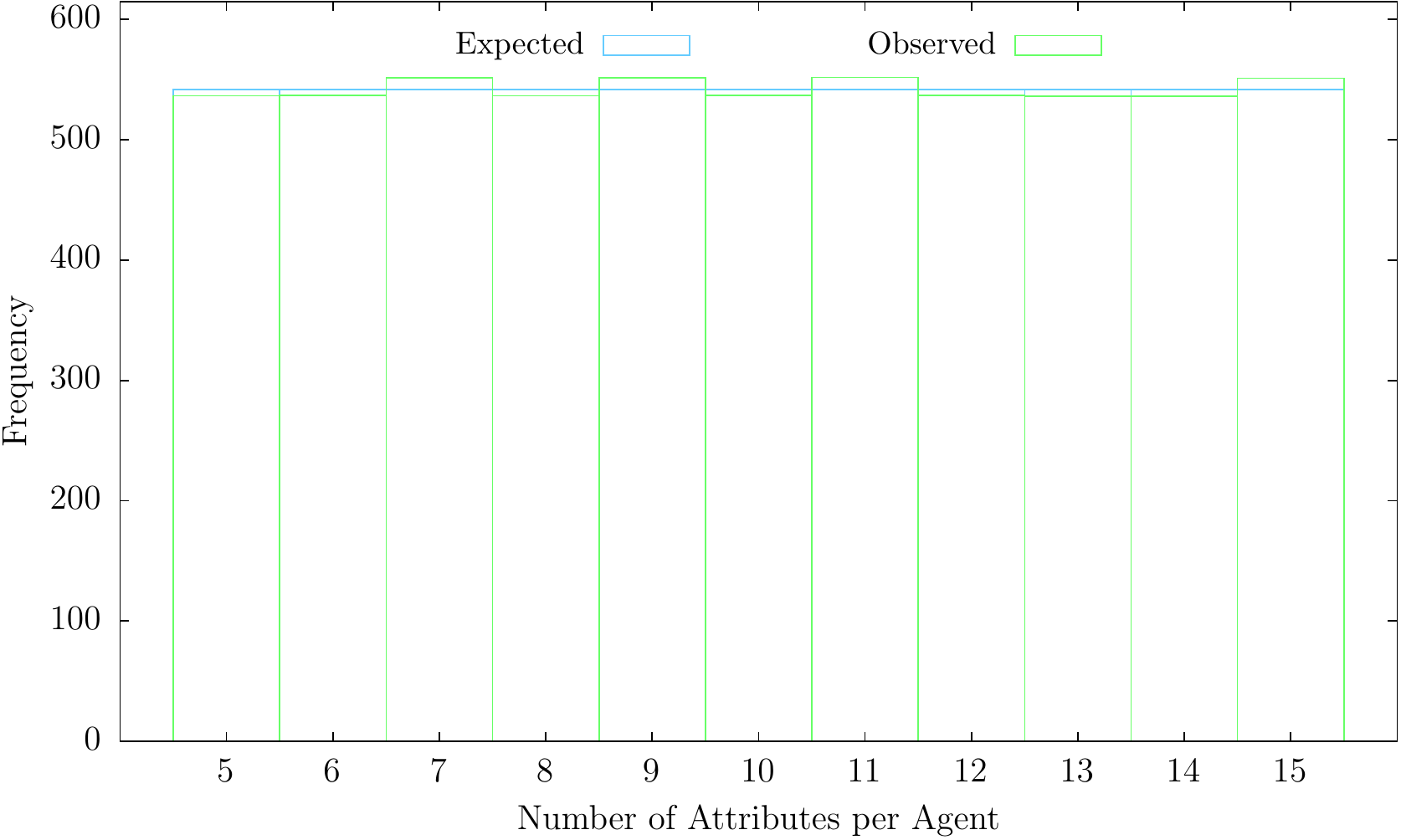}{graph}{Graph of Uniformly Distributed Agent Attribute Frequencies}{\changed{\getCap{urvunifromCap},} \added{which was confirmed by a $\chi^2$ test; with a \emph{null hypothesis} of \emph{no significant difference} and \emph{ten degrees of freedom}, the $\chi^2$ value was 1.049, below the critical 0.95 $\chi^2$ value of 3.940.}}{0mm}{!b}{0mm}{}

\tfigure{width=\textwidth}{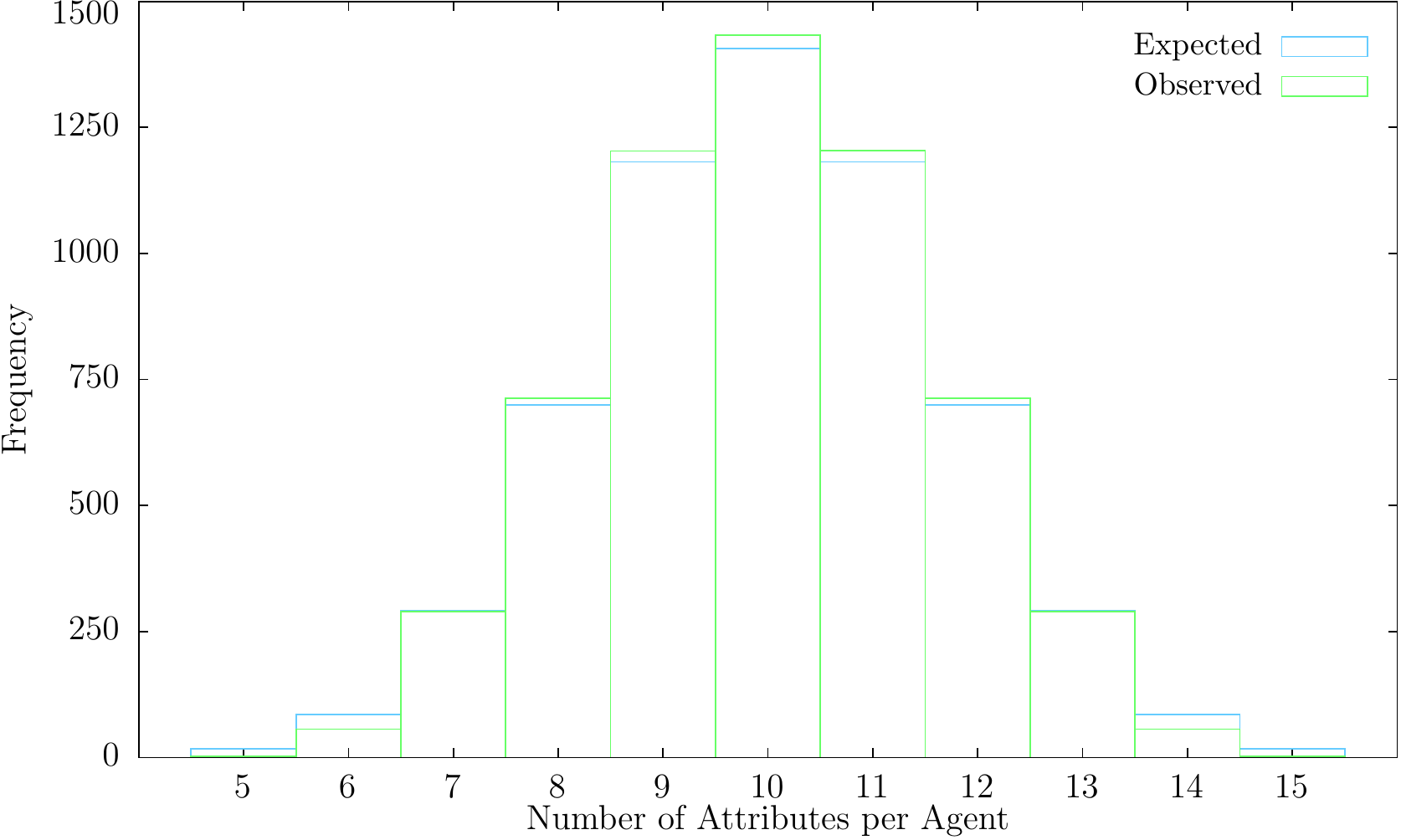}{graph}{Graph of Gaussian Distributed Agent Attribute Frequencies}{\changed{\getCap{urvgaussianCap},} \added{which led to a failed $\chi^2$ test; with a \emph{null hypothesis} of \emph{no significant difference} and \emph{ten degrees of freedom}, the $\chi^2$ value was 50.623, not below the critical 0.95 $\chi^2$ value of 3.940.}}{-3mm}{!t}{0mm}{0mm}

\tfigure{width=\textwidth}{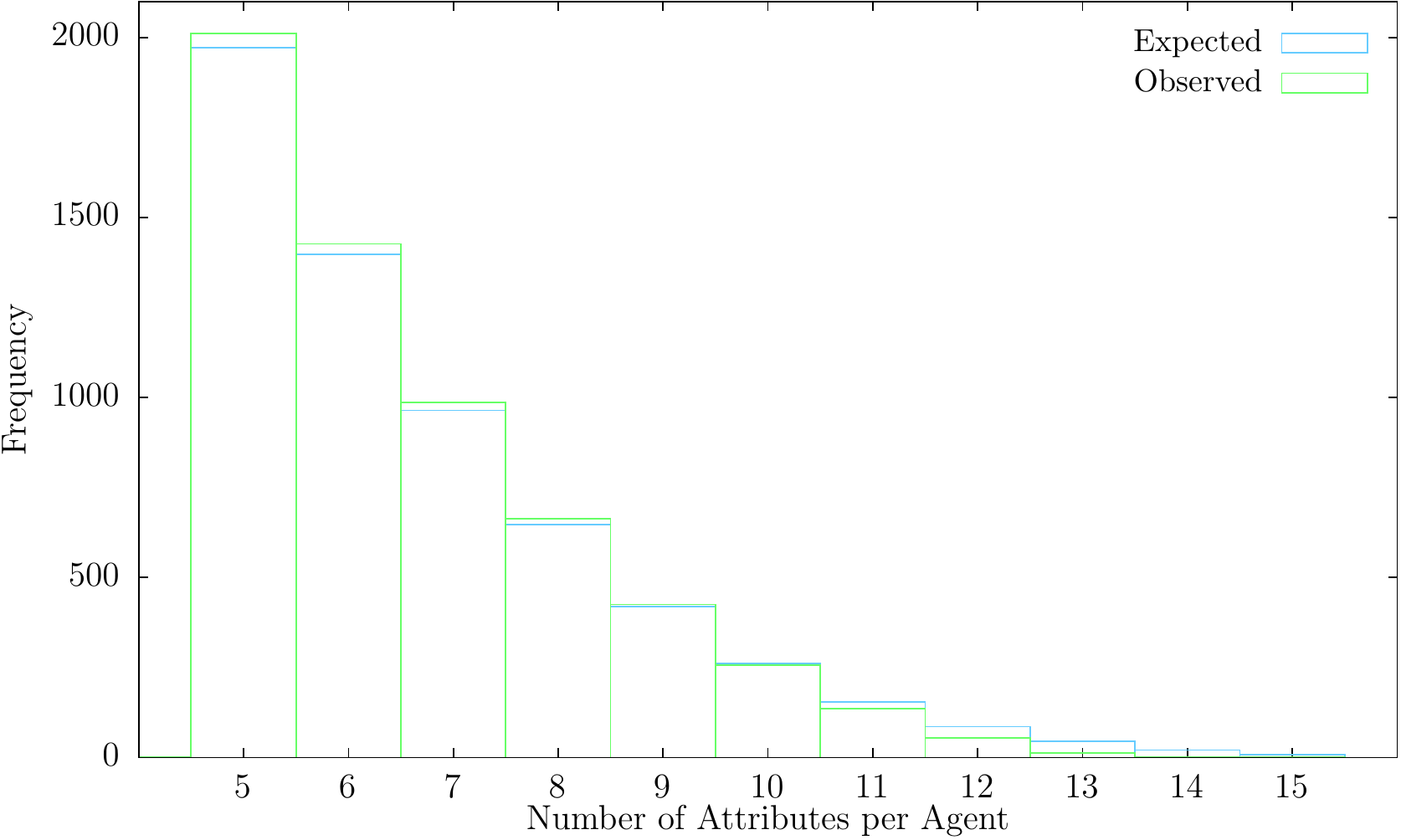}{graph}{Graph of Power Law Distributed Agent Attribute Frequencies}{\changed{\getCap{urvpowerCap},} \added{which led to a failed $\chi^2$ test; with a \emph{null hypothesis} of \emph{no significant difference} and \emph{ten degrees of freedom}, the $\chi^2$ value was 61.876, not below the critical 0.95 $\chi^2$ value of 3.940.}}{-3mm}{!h}{0mm}{-11mm}

\changed{In all of the experiments the \emph{observed} frequencies for the number of agent attributes followed the \emph{expected} frequencies, with some variation in two of the experiments.} Collectively, the experimental results confirm that the self-organised \emph{diversity} of the evolving agent populations is relative to the \emph{selection pressures} of the user base, which was confirmed statistically for most of the experiments. \changed{While the minor experimental failures, in which the Digital Ecosystem responded more slowly than in the other experiments, shows the potential to optimise the Digital Ecosystem, because the evolutionary self-organisation of an ecosystem is a slow process \citep{begon96}, even the accelerated form present in our Digital Ecosystem.}

\subsubsection{Summary}

\changed{Collectively, the experimental results confirm that the self-organised \emph{diversity} of the evolving agent populations is relative to the \emph{selection pressures} of the user base, which was confirmed statistically for most of the experiments. So, we have determined an effective understanding and quantification for the self-organised \emph{diversity} of the evolving agent populations of our Digital Ecosystem. While the minor experimental failures, in which the Digital Ecosystem responded more slowly than in the other experiments, have shown that there is potential to optimise the Digital Ecosystem, because the evolutionary self-organisation of an ecosystem is a slow process \citep{begon96}, even the accelerated form present in our Digital Ecosystem.}

\section{Conclusions}

\added{Overall an insight has been achieved into where and how self-organisation occurs in our Digital Ecosystem, and what forms this self-organisation can take and how it can be quantified. The hybrid nature of the Digital Ecosystem resulted in the most suitable definition for the self-organised \emph{complexity} coming from the biological sciences, while the most suitable definition for the self-organised \emph{stability} coming from the computer sciences. However, we were unable to use any existing definition for the self-organised \emph{diversity}, because the hybrid nature of the Digital Ecosystem makes it unique, and so we constructed our own definition based on variation relative to the user base. The (Physical) \emph{complexity} definition applies to a single point in time of the evolving agent populations, whereas the (Chli-DeWilde) \emph{stability} definition applies at the end of these instantiated evolutionary processes, while our \emph{diversity} definition applies to the optimality of the distribution of the agents within the evolving agent populations of the Digital Ecosystem. The experimental results have generally supported the hypotheses, and have provided more detail to the behaviour of the self-organising phenomena under investigation, showing some of its properties and for the self-organised \emph{diversity} has shown that there is potential for optimising the Digital Ecosystem, which we have since considered here \citep{thesis,de08th}.}

\subsection{\added{Complexity}}
\added{We investigated the self-organised \emph{complexity} of evolving agent populations through experimental simulations, for which our extended Physical Complexity was consistent with the original. We then investigated the \emph{Efficiency}, which performed as expected, confirmed by the numerical results and population visualisations matching our intuitive understanding. We then applied the \emph{Efficiency} to the determination of clusters when subjecting an evolving agent population to a multi-objective \emph{selection pressure}. The numerical results, combined with the visualisation of the multi-cluster population, confirmed the ability of the \emph{Efficiency} to act as a \emph{clustering coefficient}, not only indicating the occurrence of clustering, but also the number of clusters (for pure clusters). We also confirmed that the \emph{Efficiency $E_{c}$ for populations with clusters} was able to calculate correctly the self-organised \emph{complexity} of evolving agent populations with clusters.}

\added{We have determined an effective understanding and quantification for the self-organised \emph{complexity} of the evolving agent populations of our Digital Ecosystem. Furthermore, the understanding and techniques we have developed have applicability beyond evolving agent populations, as wide as the original Physical Complexity, which has been applied from \acs{DNA} \citep{adami2000} to simulations of self-replicating programmes \citep{adami20032}.}

\subsection{\added{Stability}}
\added{We then investigated the self-organised \emph{stability} of evolving agent populations through experimental simulations, and the results showed that there was a limit probability distribution, and that it was non-uniform. Furthermore, the reaching of the \emph{maximum macro-state} was confirmed by a visualisation matching the numerical results. We then applied our \emph{degree of instability} to determine that there was no instability under normal conditions, and then performed a \emph{stability analysis} (similar to a \emph{sensitivity analysis} \citep{cacuci2003sau}) showing the variation of the self-organised \emph{stability} under varying conditions.}

\changed{We have determined an effective understanding and quantification for the self-organised \emph{stability} of the evolving agent populations of our Digital Ecosystem. Also, our extended Chli-DeWilde stability is applicable to other \aclp{MAS} with evolutionary dynamics. Furthermore, our \emph{degree of instability} provides a definition for the \emph{level of stability}, applicable to \aclp{MAS} with or without evolutionary dynamics.}

\subsection{\added{Diversity}}

\added{We then investigated the self-organised \emph{diversity} of evolving agent populations through experimental simulations. First, varying the \emph{user request length} according to the different distributions, and testing whether the \emph{observed} frequencies of the agent-sequence length matched the \emph{expected} frequencies, which we confirmed with successful chi-squared tests. Second, varying the \emph{user request modularity} according to the different distributions, and testing whether the \emph{observed} frequencies for the number of agent attributes matched the \emph{expected} frequencies, again confirming with chi-squared tests. Under the Gaussian and Power distributions the chi-squared tests failed, most likely because the evolving agent populations were still self-organising to match the user behaviour, because at the time the Digital Ecosystem was sampled each user had placed an average of only ten requests.}

\bibliographystyle{jmr}
\bibliography{references,myRefs} 
\end{document}